\documentclass[11pt]{article}

\usepackage[a4paper, left=2.2cm, right=2.2cm, top=78pt, bottom=80pt, headheight=14pt, headsep=13pt, footskip=26pt]{geometry}

\usepackage{XCharter}
\usepackage[xcharter, vvarbb]{newtxmath}

\usepackage{amsmath,amsfonts,bm}

\def\eqref#1{equation~\ref{#1}}

\def\1{\bm{1}}

\DeclareMathAlphabet{\mathsfit}{\encodingdefault}{\sfdefault}{m}{sl}
\SetMathAlphabet{\mathsfit}{bold}{\encodingdefault}{\sfdefault}{bx}{n}

\usepackage{natbib}
\setcitestyle{authoryear,round,citesep={;},aysep={,},yysep={;}}
\usepackage{float}
\usepackage{xcolor}
\usepackage[colorlinks=true, linkcolor=arxivlink, citecolor=arxivlink, urlcolor=arxivlink]{hyperref}
\usepackage{url}
\usepackage{graphicx}
\usepackage{wrapfig}
\usepackage{booktabs}
\usepackage{array}
\usepackage{comment}
\usepackage{etoolbox}
\usepackage{fancyhdr}
\usepackage{titlesec}
\usepackage[labelfont=bf]{caption}
\usepackage[most]{tcolorbox}
\usepackage{fontawesome5}

\definecolor{arxivlink}{HTML}{2420CF}
\definecolor{arxivrule}{HTML}{808080}
\definecolor{abstractbg}{HTML}{F4F7FA}
\definecolor{abstractbar}{HTML}{4788BD}
\definecolor{abstracttext}{HTML}{26333F}
\definecolor{abstractaccent}{HTML}{C87945}

\titleformat{\section}{\fontsize{13}{16}\selectfont\bfseries}{\thesection.}{0.5em}{}
\titleformat{\subsection}{\normalsize\bfseries}{\thesubsection.}{0.5em}{}
\titleformat{\subsubsection}{\normalsize\bfseries}{\thesubsubsection.}{0.5em}{}
\titleformat{\paragraph}[runin]{\normalsize\bfseries}{}{0pt}{}
\titlespacing*{\section}{0pt}{14pt plus 4pt minus 2pt}{6pt plus 2pt}
\titlespacing*{\subsection}{0pt}{12pt plus 4pt minus 2pt}{4pt plus 2pt}
\titlespacing*{\subsubsection}{0pt}{10pt plus 3pt minus 2pt}{2pt plus 1pt}
\titlespacing*{\paragraph}{0pt}{6pt plus 2pt minus 1pt}{0.6em}

\DeclareCaptionLabelSeparator{bar}{\ \textbar\ }
\AtBeginEnvironment{table}{\setlength{\abovecaptionskip}{0pt}\setlength{\belowcaptionskip}{10pt}}
\AtBeginEnvironment{table}{\let\small\footnotesize}

\newcommand{\papertitle}{What Effort Level Should I Choose? On the Economics of Reasoning Effort in LLM Agents}
\hypersetup{pdftitle={\papertitle}, pdfauthor={Kaiser Sun, Yongchan Kwon, Zain Hasan, Shang Zhu, Kaitlyn Zhou, Federico Bianchi}}

\makeatletter
\newcommand{\arxivrule}{{\color{arxivrule}\hrule\@height 1pt\@width\headwidth}}
\newif\ifarxivfootnote
\let\arxiv@makecol\@makecol
\def\@makecol{\ifvoid\footins\arxivfootnotefalse\else\arxivfootnotetrue\fi\arxiv@makecol}
\makeatother
\renewcommand{\headrulewidth}{1pt}
\renewcommand{\footrulewidth}{1pt}
\renewcommand{\headrule}{\arxivrule\vskip-1pt}
\renewcommand{\footrule}{\ifarxivfootnote\else\vskip-\footruleskip\vskip-1pt\arxivrule\vskip\footruleskip\fi}
\renewcommand{\footnoterule}{\kern-3pt\arxivrule\kern2pt}

\fancypagestyle{arxivfirst}{%
\fancyhf{}%
\fancyhead[L]{\smash{\includegraphics[height=0.78cm]{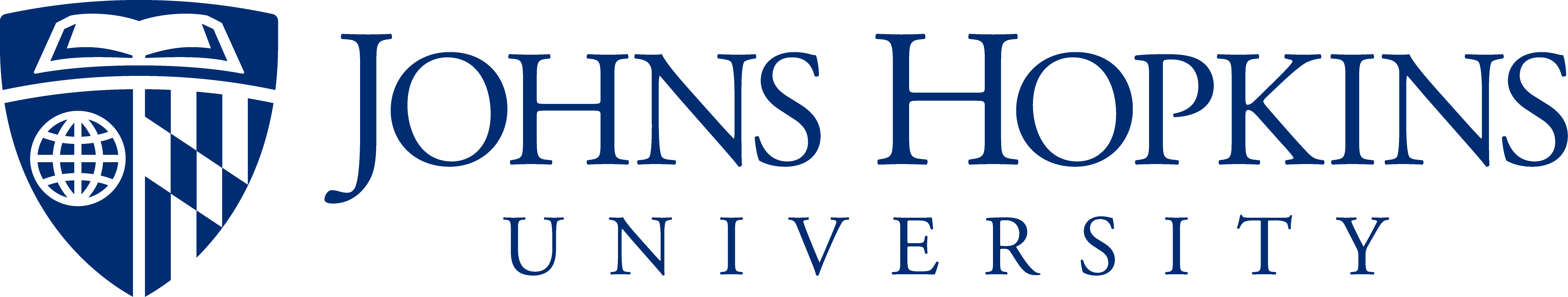}\hspace{1.4em}\includegraphics[height=0.75cm]{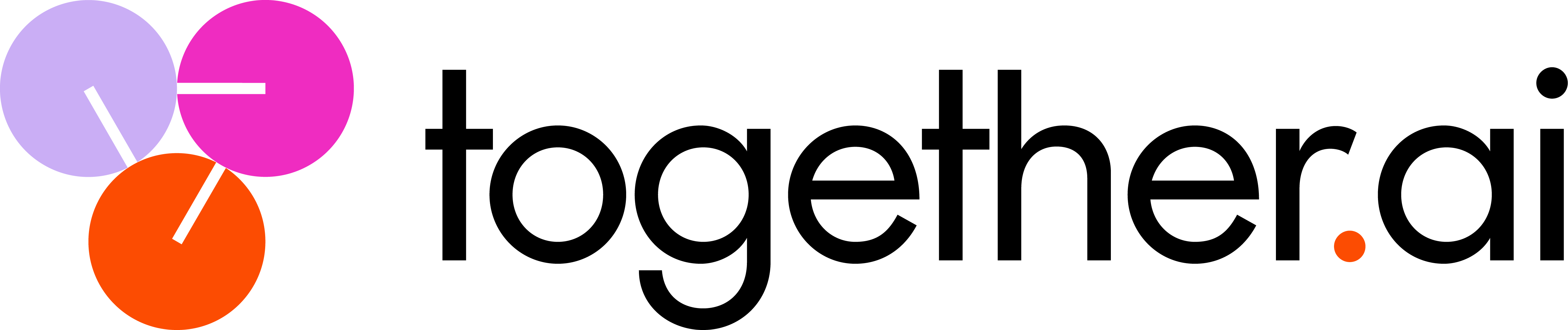}\hspace{1.4em}\includegraphics[height=0.9cm]{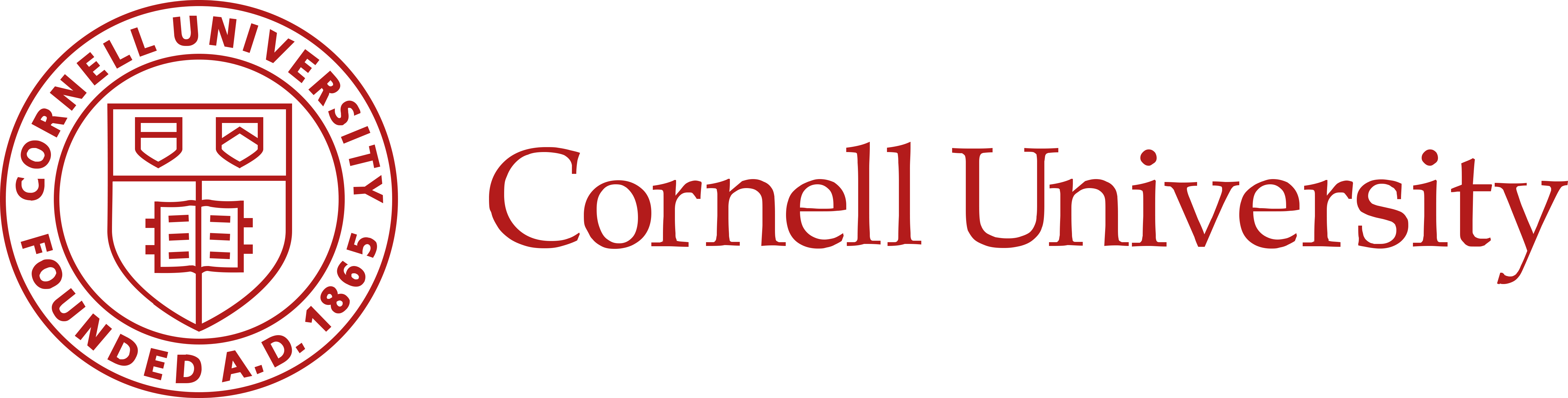}}}%
\fancyfoot[R]{\small\thepage}%
}

\newtcolorbox{abstractbox}{enhanced, sharp corners, colback=abstractbg, colframe=abstractbar, boxrule=0pt, leftrule=1.6pt, left=8pt, right=8pt, top=6pt, bottom=8pt, before skip=10pt, after skip=14pt, fontupper=\fontsize{10}{13}\selectfont\color{abstracttext}}
\newcommand{\abstractlabel}{{\color{abstractbar}\sffamily\bfseries\fontsize{9}{11}\selectfont A\kern.16em B\kern.16em S\kern.16em T\kern.16em R\kern.16em A\kern.16em C\kern.16em T}\hspace{0.7em}{\color{abstractaccent}\rule[0.3ex]{1.8em}{1pt}}}

\newcommand{\actonomy}{Act\textperiodcentered ONOMY}

\newcommand{\arxivversion}{}

\begin{document}

\thispagestyle{arxivfirst}

{\raggedright\fontsize{17.5}{22}\selectfont\bfseries What Effort Level Should I Choose?\\On the Economics of Reasoning Effort in LLM Agents\par}
\vspace{10pt}

{\raggedright\fontsize{9}{12}\selectfont\setlength{\parskip}{0pt}
{\bfseries Kaiser~Sun$^{1*\dagger}$, Yongchan~Kwon$^{2}$, Zain~Hasan$^{2}$, Shang~Zhu$^{2}$, Kaitlyn~Zhou$^{3,2}$, Federico~Bianchi$^{2\dagger}$}\par
\vspace{5pt}
$^{1}$Johns Hopkins University\enspace $^{2}$Together AI\enspace $^{3}$Cornell University\par
$^{\dagger}$Corresponding authors\par}
{\renewcommand{\thefootnote}{*}\footnotetext{Work done during an internship at Together AI.}}

\begin{abstractbox}
\abstractlabel\par
\vspace{4pt}

Generative models increasingly expose a \textit{reasoning effort} setting that lets users scale inference-time compute, trading higher cost for potentially improved performance.
Users must choose an effort level before knowing whether extra reasoning will help, and even a successful run does not reveal whether the effort purchased was necessary. Reasoning effort is therefore a \emph{credence good}: buyers observe the outcome but cannot assess the marginal value of the computation they paid for.
In this work, we measure the counterfactual effect of effort levels on both proprietary and open-weight models on four benchmarks, spanning single-call reasoning and long-horizon agentic tasks. \textit{Effort}, it turns out, does not consistently preserve tasks that lower effort solves, can overspend, and can change trajectory behavior. 
First, we find that although higher effort can raise accuracy in \textit{aggregate}, it frequently fails on tasks that lower effort solved. 
Second, on the subset of easy tasks that every effort level solves, higher effort is generally more costly, with the highest effort level costing up to 42$\times$ as much as the lowest. 
Lastly, \textit{effort} changes not only how many tokens are used, but also \emph{how} agents solve tasks: at higher effort, agents verify their own work more heavily and more often search for solutions than work through problems directly, both of which lengthen trajectories without necessarily altering the outcome. 
These findings show that naively choosing the highest level is not only cost-inefficient but also offers no guarantee of the best instance-level performance. 
A simple alternative, starting at the lowest effort and escalating only after a verified failure, matches or improves on fixed top effort across all models and benchmarks at lower cost for most proprietary models. %

\par\vspace{6pt}
{\color{black}\faGithub\enspace\textbf{Code:} \href{https://github.com/KaiserWhoLearns/ReasoningEffortLLM}{https://github.com/KaiserWhoLearns/ReasoningEffortLLM}}\par
\end{abstractbox}

\section{Introduction}\label{sec:intro}
As large language models (LLMs) are deployed more widely and applied to long-horizon agentic tasks, inference cost has become a central concern.
Many reasoning models now expose an \textit{effort} parameter that adjusts inference-time computation without changing the underlying model \citep{openai2026reasoning,anthropic2026effort}.
This effort parameter scales \textit{test-time compute}, offering better performance on hard problems without switching to a larger model or fine-tuning for the task \citep{snell2025scaling,muennighoff2025s1}.
In agentic settings, effort also changes how the model acts, including how much it plans and how many tools it calls \citep{anthropic2026effort}.
Because providers bill for input, output, and reasoning tokens alike, cost often rises with effort.

However, users must select an effort level without knowing whether the additional computation will help on their task.
This uncertainty makes reasoning effort a \emph{credence good}: buyers cannot readily assess how much of a service they need, even after observing the outcome \citep{darby1973free,dulleck2006doctors}.
A successful run shows that the task was solved, but not that the purchased effort was necessary, and rerunning at another level to find out is rarely worth the cost.
Users therefore pay for reasoning whose contribution they cannot observe.

In this paper, we study the following counterfactual question: had the user chosen a different, typically lower, effort level, how would the outcome and its cost have changed? Answering this matters because effort is now the main lever users have over inference spend: as agents run for hundreds of calls and reprocess growing contexts, a single effort setting is multiplied across every step of every trajectory, and it is selected once, by intuition, with no feedback on whether it was needed.
We measure the performance, cost, and behavioral consequences of reasoning effort for proprietary and open-weight models on four benchmarks.
Within each model--benchmark configuration, we vary effort while holding tasks, prompts, and the agentic harness fixed where applicable.
In our experiments, higher effort does not consistently preserve lower-effort successes, yet it often leads to higher costs.
We characterize this mismatch from four angles and show how allocating effort per task reduces it:
\begin{enumerate}
\item \textbf{Aggregate performance hides instances lost at higher effort (\S\ref{sec:perf}).}
Higher effort often solves more tasks in aggregate, but it may lose on tasks that lower effort can solve.

\item \textbf{Higher effort overspends on tasks that lower effort already solves (\S\ref{sec:cost}).}
On tasks solvable at every effort level, models' mean costs reach up to $42\times$ their \texttt{lowest}-effort cost, driven by both more computation per call and longer trajectories that repeatedly reprocess a growing context.
Disabling reasoning does not reliably reduce cost either, because agents compensate with longer trajectories.

\item \textbf{Different efforts produce different solution strategies (\S\ref{sec:behavior}).}
Higher effort shifts agents toward heavier self-verification and shortcut behaviors such as retrieving existing answers, which are not suitable for all tasks.

\item \textbf{Effort should be allocated per task (\S\ref{sec:reco}).}
Models cannot reliably choose effort from the task description alone, but \emph{na\"ive laddering}, which escalates from the lowest to the highest effort only after a failed correctness check, matches or improves the success of fixed top effort across all models and benchmarks, while maintaining a lower median inference cost when considering the user waiting time.
\end{enumerate}
Taken together, higher reasoning effort is not necessarily a reliable purchase: its benefits depend on the task, while its costs rise regardless, a pattern that persists across repeated runs.
Neither maximal effort nor disabled reasoning is a safe default, so effort should be dynamic rather than fixed in advance, and evaluations that count accuracy and tokens alone may misjudge which configuration is actually better.
As long as users can observe neither the reasoning content they receive nor whether it mattered, they remain credence-good buyers. We argue that reasoning charges should be verifiable, so that paying for more thought becomes an informed choice rather than an act of faith.

\section{Experimental setup}
\label{sec:setup}

We evaluate how reasoning effort affects task success, complete execution cost, and agent behavior across two single-call LLM benchmarks and two agentic benchmarks.
Within each model--benchmark configuration, we vary the reasoning-effort setting while holding the tasks, prompts, and, where applicable, the agentic harness fixed.
This design allows us to compare effort settings within a configuration and examine how their effects differ across models and execution environments.

\paragraph{Benchmarks and execution environments.}
AIME 2026 \citep{dekoninck2026matharena} evaluates mathematical problem solving, while GPQA Diamond \citep{rein2023gpqa} evaluates graduate-level science question answering.
Both use a single inference call per attempt.
Terminal-Bench v2.1 \citep{merrill2026terminal} evaluates agents on tasks in a terminal environment using Terminus-2, and SWE-Bench Pro \citep{deng2025swe} evaluates repository-level software engineering using \texttt{mini-swe-agent}.
These agentic benchmarks involve multi-turn interactions, in which effort can affect tool use, verification, and stopping behavior as well as computation within each call.
Processing the accumulated interaction history also makes input-token consumption sensitive to trajectory length.
Since the original SWE-Bench Pro containers contain the repository history that can expose the reference fix, we evaluate a corrected variant that removes git history and restricts network access to allowlisted package registries (\S\ref{sec:behavior}).
We report original and corrected SWE-Bench Pro separately: the main instances-gained and always-solved cost figures use the original environment, while the laddering summary uses the corrected environment.\footnote{Appendix Table~\ref{tab:app_perf}, Appendix Figures~\ref{fig:app_instances_gained}, and~\ref{fig:app-swebenchpro-corrected-cost} reports the same experimental results in the main section for the alternative environment.}
Appendix~\ref{sec:appendix-benchmarks} summarizes evaluation sizes and execution settings in Table~\ref{tab:benchmarks}. 

\paragraph{Models and effort levels.}
We study proprietary models GPT-5.6-sol, GPT-5.6-terra, GPT-5.6-luna \citep{openai2026gpt56}, and Claude Sonnet 5 \citep{anthropic2026claudesonnet5}, alongside the open-weight models GLM-5.2 \citep{zai2026glm52}, Inkling \citep{thinkingmachines2026inkling}, Kimi K3 \citep{team2026kimi}, and DeepSeek-V4-Flash-0731 \citep{xu2026deepseek}.
Each model has different official reasoning effort settings, as listed in Table~\ref{tab:inference}.
We query the proprietary models through their providers' APIs and the open-weight models primarily through Together AI, set reasoning effort through each provider's native control, and leave sampling parameters at the provider or harness defaults (Appendix~\ref{sec:appendix-inference}).

\paragraph{Measurement and cost accounting.}
We record task outcomes and capture token usage for every inference call, aggregating usage over the complete execution for agentic tasks.
For AIME and GPQA, we run $k$ independent attempts per task at each effort level, with $k=10$ for AIME and $k=4$ for GPQA, matching the commonly reported settings.
We use pass@k to determine whether a task is solved and mean@k to report aggregate performance.
Due to budget constraints, each agentic task is executed once per effort level.\footnote{This is a limitation we accept to cover eight models and their model-specific effort settings.
While no single run is conclusive on its own, the observed reversals and cost increases recur across models and benchmarks.}
To provide additional evidence that quantifies the uncertainty introduced by this single-run protocol, we analyze four independent runs' trajectories on DeepSWE \citep{huang2026deepswe} in Appendix~\ref{sec:appendix-repeated-runs}.
Cost accounting includes input and output tokens, including billed reasoning tokens.
We convert usage to dollars using a rate table published by each model provider.
Keeping token measurements separate from monetary estimates allows costs to be computed with the most recent pricing data available at the time of submission.

\section{Findings}
\label{sec:findings}
Reasoning effort is exposed as a single ordered dial, with the implicit assumption that by increasing test-time compute, performance will scale accordingly. Here, we test this assumption from the perspectives of accuracy, cost efficiency, and behavior. At a high level, we find that higher effort does not preserve lower-effort successes (\S\ref{sec:perf}), higher effort overspends on tasks that lower effort already solves (\S\ref{sec:cost}), and higher efforts change how agents solve tasks rather than simply deliberating longer (\S\ref{sec:behavior}).
Together, these findings illustrate the limitations of fixed effort levels and motivate future work on dynamically allocating effort by task (\S\ref{sec:reco}).

\subsection{Aggregate performance hides instances lost at higher effort}
\label{sec:perf}

\begin{figure}[t]
    \centering
    \includegraphics[width=\linewidth]{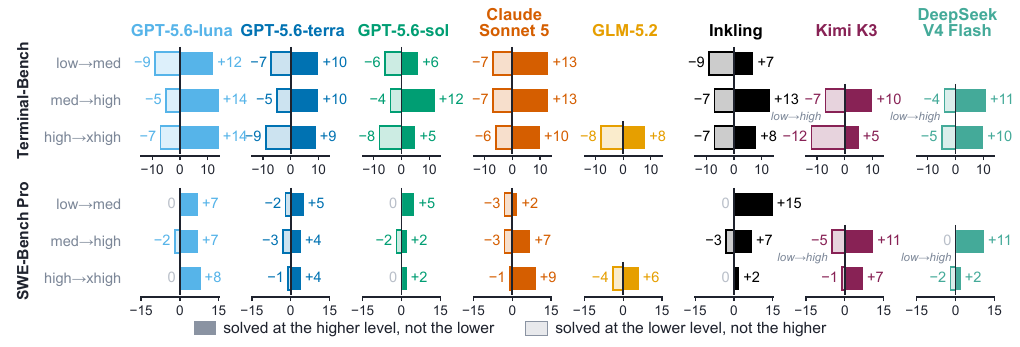}
    \caption{
    Instances gained and lost between adjacent reasoning-effort settings on the agentic benchmarks.
    Higher effort can solve additional instances while losing some lower-effort successes.
    AIME, GPQA, and a corrected version of SWE-Bench Pro are included in Appendix Figure~\ref{fig:app_instances_gained}.
    }
    \vspace{-\baselineskip}
    \label{fig:instances_gained}

\end{figure}

Reasoning effort settings are presented to users as an ordered sequence, typically spanning \texttt{None}, \texttt{low}, \texttt{medium}, \texttt{high}, \texttt{extra high}, and \texttt{max}.
Let $s_e$ denote the set of instances a model solves at effort level $e$, and let $e_1 < e_2$ order two such levels.
One might read this ordering as that raising effort only ever \emph{adds} solutions: the solved sets are nested, $s_{e_1} \subseteq s_{e_2}$, and in particular $|s_{e_1}| \le |s_{e_2}|$.
Higher effort often does solve more instances, so $|s_{e_1}| < |s_{e_2}|$ frequently holds.
The nesting, however, does not always hold: models \textbf{fail} on tasks they had solved at lower effort, so the \emph{regression set} $s_{e_1} \setminus s_{e_2}$ is nonempty, and we call these observed reversals "regressions", not established decreases in task-specific success probabilities.
These reversals can arise from execution variability even when higher effort does not reduce a task's success probability.
Regardless of their cause, aggregate performance alone does not reveal which instances succeed at lower effort and fail at higher effort.

Take Inkling on GPQA as an example. Raising effort from \texttt{low} to \texttt{medium} gives a net gain of $|s_{\texttt{medium}}| - |s_{\texttt{low}}| = 16$ questions. 
But this figure is the difference between two \textbf{disjoint} sets, and it hides the regression set $s_{\texttt{low}} \setminus s_{\texttt{medium}}$. It turns out that at \texttt{medium}, Inkling failed all four attempts on five questions that \texttt{low} had solved, $|s_{\texttt{low}} \setminus s_{\texttt{medium}}| = 5$.  
The aggregated gain (+16 tasks), which is frequently what is reported in the evaluation benchmarks, obscures the questions that the higher level fails to solve (Appendix Figure~\ref{fig:app_instances_gained}).

Such regressions, although less apparent on non-agentic benchmarks, occur across both agentic benchmarks (Figure~\ref{fig:instances_gained}).
On Terminal-Bench, every one of the eight models we evaluated has a nonempty regression set between its lowest and highest effort levels.
For GPT-5.6-sol, moving from \texttt{high} to \texttt{xhigh} loses eight tasks while newly solving only five, so the solved count falls from 76 to 73 of 89 tasks.
Aggregate counts $|s_e|$ can therefore make higher effort look like a strict improvement while hiding the lower-effort successes it gives up in $s_{e_1} \setminus s_{e_2}$.
On the agentic benchmarks, many solved sets are not nested: higher effort can lose lower-effort successes even when aggregate performance improves.
\footnote{Because each agentic task runs once, some reversals reflect sampling noise.
Our four-rollout ablation study (Appendix~\ref{sec:appendix-repeated-runs}) shows that such behavior persists at rates comparable to chance, so we report them as observed reversals rather than systematic decreases.} 
In expectation, higher effort helps, but no user receives the expectation. 
A user receives one run, and on one run the higher setting's solved set need not be a superset of the lower's; the user cannot tell which successes were forfeited (§\ref{sec:reco}).

\subsection{Higher effort overspends on tasks that lower effort already solves}
\label{sec:cost}

Beyond regressing on tasks solved at a lower effort level, models exhibit a separate inefficiency: higher-effort executions can cost more than successful lower-effort executions of the same task.
One might hope that a model told to spend heavy effort on an easy problem would solve it quickly and stop once there was nothing left to deliberate over. Unfortunately, we find that language model agents show no such tendency.
Under higher effort, token usage on easy instances does not contract toward the lower-effort cost; instead, higher effort consumes far more tokens on tasks that a lower level had already solved more cheaply.
We observe that \textit{effort} behaves more like a fixed multiplier on computation, rather than an upper-bound budget that can be calibrated to task difficulty.

\begin{figure}[t]
    \centering
    \vspace{-\baselineskip}
    \includegraphics[width=\linewidth]{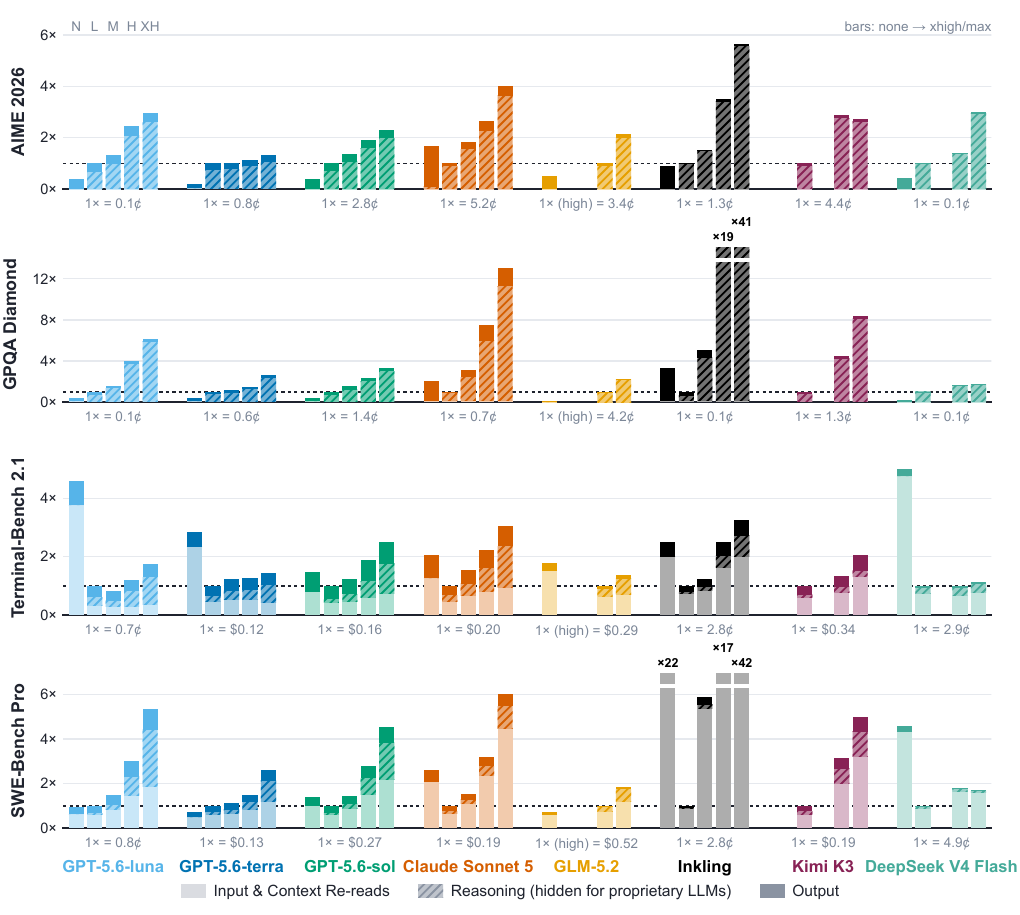}
    \vspace{-23pt}
    \caption{API cost of one task the model solves at every effort level, including \texttt{none} if supported, relative to the cost at \texttt{low} (dashed line; GLM-5.2 is anchored on \texttt{high}, its lowest reasoning level).
    Kimi K3 has no \texttt{none} setting.
    Each bar is split into input, hidden reasoning, and visible output tokens.
    The always-solved set is fixed per model and benchmark, so bars compare prices within a group and only vary across groups; the footer gives the dollar value of $1\times$.
    Bars taller than the axis are clipped and labelled.}
    \vspace{-\baselineskip}
    \vspace{-5pt}
    \label{fig:solvable_cost}
\end{figure}

To isolate this effect, for each model and benchmark, we restrict the comparison to the intersection of tasks solved at every effort level, $s_{all}$.
On AIME and GPQA, membership in $s_{all}$ uses pass@k, while each task's cost is averaged over all $k$ attempts at that effort level.
Conditioning on this always-solved subset compares spending across effort levels without conflating cost differences with gains from solving additional tasks.
On $s_{all}$, \textbf{higher reasoning effort substantially increases inference cost}.
On Terminal-Bench, increasing effort from \texttt{low} to \texttt{xhigh} raises mean cost by factors of $2.51$ for GPT-5.6-sol and $3.04$ for Claude Sonnet 5.
On SWE-Bench Pro, GPT-5.6-sol's mean cost rises from \$0.27 to \$1.24 over the always-solved tasks ($4.54\times$), while GPT-5.6-luna's rises by $5.34\times$.
The increase is not confined to reasoning tokens: for GPT-5.6-sol on SWE-Bench Pro, mean input-token charges rise from \$0.17 to \$0.60 as mean trajectory length increases from 14.7 to 28.4 model calls.
Additional effort therefore raises cost through two mechanisms: more computation within individual calls and longer trajectories that repeatedly process an expanding interaction history.
Figure~\ref{fig:solvable_cost} reports mean inference cost relative to the lowest reasoning effort available for each model (\texttt{low}, or \texttt{high} for GLM-5.2, whose reasoning-enabled settings begin at that level).

Models configured with high effort do not recover the efficient route they took at lower effort, and spending scales with effort even on tasks that every level solves.
On these tasks, a low effort is sufficient: higher effort buys no additional success, and disabling reasoning shifts the cost into longer trajectories.

\textbf{"None" effort is not zero additional cost.}
Our results also highlight that using effort None is not a silver bullet that would eliminate costs. We again use $s_{all}$, which includes tasks solved at all effort levels, including when the level is set to \texttt{None}. The label of \texttt{None} reasoning suggests that it might be among the cheapest settings, as it eliminates the additional test-time compute. However, to our surprise, we find that although \texttt{None} may not generate more reasoning tokens, it can still spend more money and time than all the other effort levels through increased input-token consumption.
On $s_{all}$, GPT-5.6-luna averages 21.5 model calls at \texttt{none}, compared with 5.4 at \texttt{low}, and costs $4.58\times$ as much.
At \texttt{None}, the input-token bill increases from \$0.0021 to \$0.0258, outweighing the removal of reasoning-token charges.
Spending less on reasoning within each call can thus be offset by spending more on the trajectory as a whole.

\subsection{Different efforts produce different solution strategies}
\label{sec:behavior}

The regressions in \S\ref{sec:perf} and the overspending in \S\ref{sec:cost} suggest that effort levels do not simply run the same procedure for longer.
We examine fine-grained agent trajectories to study what changes.
Higher effort shifts agents toward retrieving existing answers and toward heavier self-verification, and neither shift reliably improves outcomes.

\paragraph{Higher effort retrieves reference fixes more often.}
In the original SWE-Bench Pro environment, the repository history can expose the gold commit that solves the task, allowing an agent to retrieve an existing solution rather than derive one from the task description and code.
GPT-5.6-sol reads the reference-fix commit on 15.9\% of tasks at \texttt{low}, compared with 75.0\% at \texttt{xhigh} (Figure~\ref{fig:swebenchpro-reference}).
Part of the apparent improvement with effort may reflect benchmark leakage rather than genuine problem solving.

To separate this retrieval strategy from problem solving, we follow \citet{jain2026reward} and create \emph{SWE-Bench Pro-Corrected}, which removes the task repository's original git history while preserving the starting file tree and working-state changes.
We also restrict network access to allowlisted package registries to limit retrieval of upstream fixes, while retaining the same tasks and evaluator.
In this corrected environment, GPT-5.6-sol still attempts online answer retrieval on 9.1\% of tasks at \texttt{low} and 68.2\% at \texttt{xhigh} even though the restrictions reject these attempts.
At \texttt{xhigh}, GPT-5.6-sol's solve rate falls from 77.3\% in the original environment to 59.1\% in the corrected environment.
All eight models have lower solve rates in the corrected environment at their highest effort setting (sign test across models, $p = 0.008$), and the decrease is individually significant for five of them (Appendix Table~\ref{tab:swebenchpro-corrected-perf}).
Lower and higher effort levels exhibit different solution strategies, and the high-effort could benefit from a strategy that low effort rarely uses.

\ifdefined\arxivversion\begin{figure}[t]\else\begin{figure}[H]\fi
    \centering
    \vspace{-5pt}
    \begin{minipage}[t]{0.38\linewidth}
        \centering
        \includegraphics[width=\linewidth]{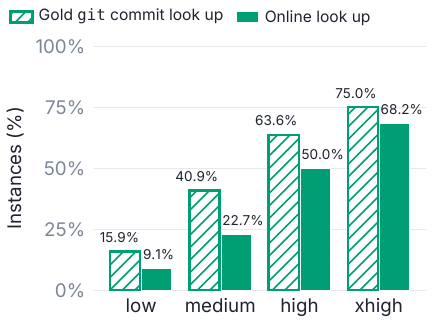}
        \small (a) Reference-solution lookup
    \end{minipage}\hspace{0.02\linewidth}
    \begin{minipage}[t]{0.38\linewidth}
        \centering
        \includegraphics[width=\linewidth]{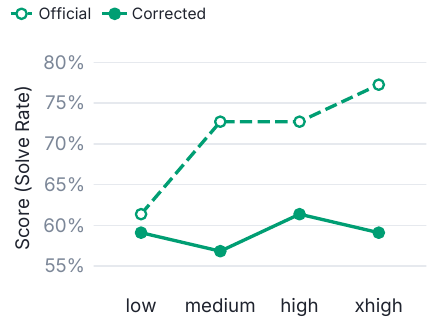}
        \small (b) Performance
    \end{minipage}
    \ifdefined\arxivversion\else\vspace{-5pt}\fi
    \caption{GPT-5.6-sol on matched SWE-Bench Pro instances across reasoning effort levels.
    Left: reference-fix reads from official git history and online lookup attempts in the corrected environment, regardless of retrieval success.     Right: instances resolved.
    Corrected removes the original git history and restricts network access to allowlisted package registries.}
    \label{fig:swebenchpro-reference}
    \ifdefined\arxivversion\else
    \vspace{-\baselineskip}
    \vspace{-10pt}
    \fi
\end{figure}

\begin{figure}
    \centering
    \includegraphics[width=\linewidth]{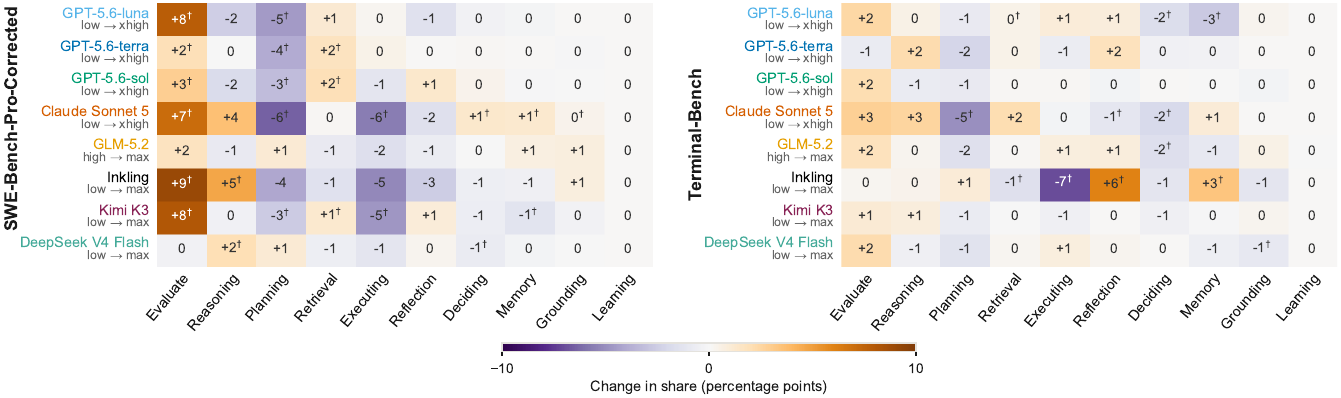}
    \vspace{-20pt}
    \caption{Changes in the mean per-trajectory share of annotated prose assigned to each \actonomy{} behavior group, in percentage points, from the lowest to the highest annotated reasoning-enabled setting on matched tasks.
    A dagger ($\dagger$) marks a change whose unadjusted 90\% paired task-bootstrap confidence interval excludes zero (10,000 resamples); markers use unrounded estimates.}
    \label{fig:actonomy-behavior-overview}
    \vspace{-17pt}
\end{figure}

\paragraph{Higher effort often leads to more verification behavior.}
We annotate the agent trajectories with the \actonomy{} taxonomy \citep{gao2026interpret}, a taxonomy of runtime agent behavior developed through grounded theory, to systematically examine what additional effort buys and inspect the corresponding actions and evaluator outputs.
It labels steps of a trajectory (e.g., a tool call or a statement in the model's output) with a category from a three-level hierarchy of 10 top-level actions (e.g., \emph{Evaluation}, \emph{Learning}), 46 subactions, and 120 fine-grained leaf categories.
We use Claude Sonnet 5 with the same annotation setting recommended by \citet{gao2026interpret}.
We use all 30 shared Terminal-Bench tasks and randomly sample 30 corrected SWE-Bench Pro instances, irrespective of outcome.
Comparing the lowest and highest annotated reasoning-enabled settings requires annotations at both endpoints, yielding 468 model--task pairs.

Figure~\ref{fig:actonomy-behavior-overview} shows the mean percentage change across ten behavior groups when switching between effort levels, with evaluation actions showing the largest change.
To examine this in more detail, we count annotations in its three leaf categories: \emph{Plan verification step}, \emph{Run verification command}, and \emph{Confirm actual == expected}, which measure evaluation/verification-related behavior among the \actonomy{} categories.
On corrected SWE-Bench Pro, all eight paired models show higher mean verification counts both on tasks passed at both endpoints and on tasks failed at both endpoints.
On Terminal-Bench, Claude Sonnet 5's mean counts increase from 6.6 to 12.3 on tasks passed at both endpoints and from 4.3 to 5.5 on tasks failed at both endpoints.
Higher effort thus devotes more attention to verification, but the increases in both outcome groups show that this additional work need not change the final result.

\paragraph{Self-verification can miss what the external evaluator checks.}
Individual trajectories show how additional checking can coexist with the regressions in \S\ref{sec:perf}.
In Terminal-Bench's \texttt{dna-insert}, GPT-5.6-sol's highest-effort run reports that its output satisfies a numerical acceptance constraint, but the evaluation verifier rejects that constraint, and the low-effort run passes.
For GLM-5.2 on \texttt{sanitize-git-repo}, the maximum-effort run contains more verification-related annotations but also rewrites repository history, removing a commit required by the external preservation check, while its lower-effort run passes.
More verification does not guarantee that the final artifact satisfies the evaluator's requirements.

Effort levels therefore differ in behavior and solution strategy, not merely in the amount of computation.
Higher effort favors retrieval and extensive self-checking, while lower effort often reaches a passing solution directly and cheaply.
Neither strategy dominates across tasks, so neither a low nor a high fixed setting is uniformly best.

\subsection{Effort should be allocated per task}
\label{sec:reco}

\begin{figure}[t]
    \centering
    
    \includegraphics[width=\linewidth]{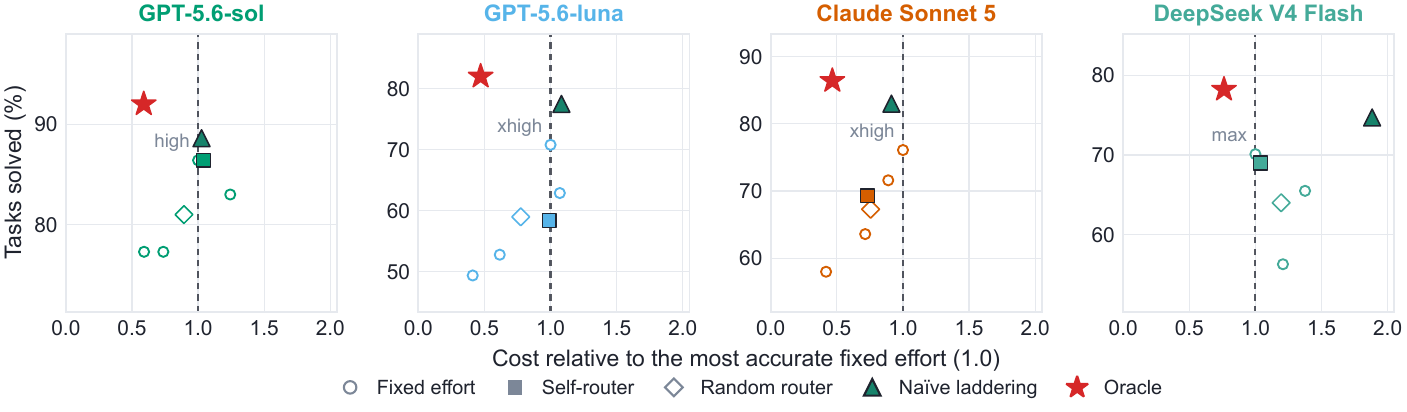}
    \vspace{-20pt}
    \caption{Self-routing and na\"ive laddering compared with fixed efforts, random allocation, and the oracle on Terminal-Bench 2.1; Appendix~\ref{sec:appendix-self-routing} shows the other models with saved self-routing comparisons.
    Each panel shows success against total API cost, normalized by the cost of that model's fixed effort with the highest performance (dashed line, effort level noted in gray text).
    Na\"ive laddering (green triangles) starts at the lowest available reasoning effort and retries at the highest effort after an externally verified failure, charging both executions when it retries.
    All policies in each panel use the same task cohort.
    Lower cost and higher success are preferable.}
    \label{fig:self_routing}
    \vspace{-\baselineskip}
    
\end{figure}

Taken together, the three findings suggest that a single fixed effort level is not optimal for LLM agents.
Higher effort, although it usually yields higher aggregate performance, loses tasks that lower effort solves (\S\ref{sec:perf}), overspends where lower effort suffices (\S\ref{sec:cost}), and pursues strategies whose value depends on the task (\S\ref{sec:behavior}).
A policy that selects effort per task could keep the successes of every level while paying for high effort only where it helps.
We quantify this opportunity with an oracle router, then evaluate two deployable policies: self-routing and na\"ive laddering (Figure~\ref{fig:self_routing}).
The oracle router selects the cheapest recorded successful effort for each task, or the cheapest recorded effort when every setting fails.
It uses outcomes and costs known only after execution and is therefore a theoretical bound rather than a deployable policy.

\paragraph{Self-routing: selecting effort before execution remains difficult.}
We first let the model choose an effort level from the task description on its own, without correctness feedback from an initial or previous attempt.
Self-routing offers no consistent cost--accuracy advantage over random allocation or na\"ive laddering, described in the following paragraph. 

\paragraph{Na\"ive laddering: outcome feedback as a stopgap.}
Na\"ive laddering starts at the lowest available reasoning effort and retries at the highest effort only after a failed correctness check, preserving initial successes and charging both attempts when escalation occurs (Appendix~\ref{sec:appendix-laddering}).
Because it keeps every initial success, laddering recovers the lower-effort successes that fixed high effort loses and avoids high-effort spending on tasks that lower effort already solves.

We find that na\"ive laddering matches or improves success over fixed top effort across all evaluated models and benchmarks, while reducing inference cost for every model on AIME and GPQA and seven out of eight models on the corrected SWE-Bench Pro.
Relative to always using each model's highest available effort setting, median inference-cost savings across models are 25.3\% on AIME, 41.5\% on GPQA, and 15.5\% on corrected SWE-Bench Pro, including the cost of both attempts when laddering retries.
On Terminal-Bench, all eight models improve success, with cost reductions for GPT-5.6-sol, GPT-5.6-terra, Claude Sonnet 5, and Kimi K3; the remaining four models gain success at increased cost.
For GPT-5.6-sol, success rises from 83.0\% to 88.6\% while cost falls by 17.4\%; for Claude Sonnet 5, success rises from 76.1\% to 83.0\% while cost falls by 8.7\%.
For Kimi K3, success rises from 73.0\% to 84.3\% while cost falls by 30.5\%.
Laddering is nonetheless a stopgap rather than a solution.
It requires a reliable external correctness check, which many deployments lack, and it pays for a full second execution on every escalation.
Its value depends on the quality of that feedback and the complete cost of retries, rather than on an assumption that higher effort is always worth buying in advance.

\begin{wrapfigure}{r}{0.45\linewidth}
    \vspace{-\baselineskip}
    \centering
    \includegraphics[width=\linewidth]{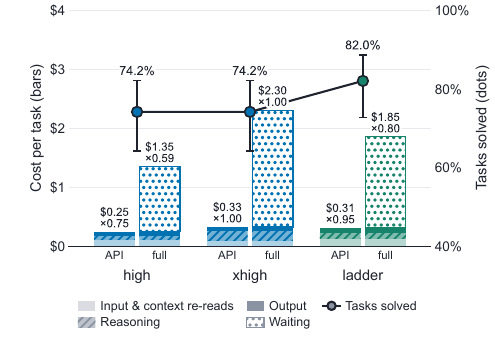}
    \vspace{-17pt}
    \caption{GPT-5.6-terra on Terminal-Bench 2.1 at fixed \texttt{high}, fixed \texttt{xhigh}, and na\"ive laddering (green).
    Bars show mean cost per task (left axis) as the API bill and as full cost (Equation~\ref{eq:full-cost}), labelled with their ratio to \texttt{xhigh}; dots show tasks solved (right axis) with 95\% Wilson intervals.}
    \label{fig:effort_pair}
    \vspace{-\baselineskip}
    \vspace{-4pt}
\end{wrapfigure}

\paragraph{Counting user waiting time, laddering saves more.}
The API bill captures inference charges but does not capture the opportunity cost of waiting while the model generates responses. Because reasoning effort can affect both inference charges and generation time, the API bill alone may misrepresent the relative cost of different effort levels from a user's perspective. To account for this additional cost, we define the estimated total cost $C_{\mathrm{full}}(i,e;m)$ for task $i$, effort level $e$, and model $m$:
\begin{equation}
    C_{\mathrm{full}}(i,e;m)
    := C_{\mathrm{API}}(i,e;m)
    + \frac{w\,N_{\mathrm{out}}(i,e;m)}{3600\,r_m},
    \label{eq:full-cost}
\end{equation}
where $N_{\mathrm{out}}$ counts all generated tokens, including reasoning tokens, and $r_m$ is the model's output throughput in tokens per second. The term $w = \$33.54$ is the mean U.S. hourly wage (Appendix~\ref{sec:appendix-full-cost}).\footnote{We use the U.S. all-occupations mean rather than a developer wage to avoid assuming who the user is. Lower wages elsewhere would shrink the waiting term but do not change the direction of the comparisons.}

Our estimated total costs make the median cost advantage of na\"ive laddering more pronounced.
For instance, on Terminal-Bench with GPT-5.6-terra, the na\"ive laddering costs 19.5\% less than fixed \texttt{xhigh} in total cost $C_{\mathrm{full}}(i,e;m)$, compared with only 4.8\% less when considering API bill alone. 
One reason is that API bills do not directly track latency. Na\"ive laddering may incur additional API cost because caching is not shared across models. In other words, it can still generate fewer tokens overall, thereby reducing waiting time. 
Because waiting time accounts for over 80\% of the total cost at every setting in Figure~\ref{fig:effort_pair}, this latency reduction can translate into a much larger cost advantage than the API bill alone suggests. More broadly, waiting time is the dominant component of our estimated cost of running LLM agents more generally (Appendix~\ref{sec:appendix-full-cost}). 
This makes latency-aware cost accounting particularly important for agentic systems, where long-running trajectories can make waiting costs substantially larger than what API bill alone captures.

\section{Related Work}

\paragraph{Test-time scaling and reasoning effort.}
Allocating more computation at inference time improves performance on hard problems \citep{snell2025scaling,muennighoff2025s1}, and providers now expose this as an ordinal reasoning effort setting \citep{openai2026reasoning,anthropic2026effort}. 
Their implementation varies across models and sometimes is only partially disclosed.
gpt-oss is trained to follow the effort keywords in the system prompt \citep{agarwal2025gpt}; 
Qwen3 enforces thinking budgets by interrupting reasoning at inference time \citep{yang2025qwen3}; 
DeepSeek-V4 distills specialists trained with different length penalties and context windows \citep{xu2026deepseek}; 
Kimi K3 penalizes trajectories exceeding per-problem token budgets, including reasoning and tool-call arguments in agentic tasks \citep{team2026kimi}; and Inkling conditions its reinforcement learning stage on a continuous effort value with a per-token cost \citep{thinkingmachines2026inkling}. 
Related methods train models to satisfy explicit length constraints \citep{aggarwal2025l1}. 
The heterogeneous mechanisms bring two effects:
First, reasoning effort, although identically named across settings, arises from different underlying mechanisms. 
Second, because the disclosed recipes induce effort largely through length-dependent rewards, nothing in training guarantees that a higher setting dominates a lower one on individual instances. 
Prior work documents overthinking on easy problems \citep{pmlr-v267-chen25bx} and in agentic software engineering \citep{cuadron2025danger}. 
We examine the performance, cost, and behavioral consequences of different reasoning efforts.

\paragraph{Cost-aware evaluation and token economics.}
\citet{kapoor2024ai} advocate cost-controlled agent evaluation using accuracy--cost Pareto frontiers. 
The Holistic Agent Leaderboard finds that higher reasoning effort reduced accuracy in the majority of its runs \citep{kapoor2026holistic}, and \citet{erol2026cost} measure the expected cost of a correct solution relative to a human-expert baseline. 
In agentic coding, input tokens rather than output tokens dominate consumption \citep{bai2026ai}; failed attempts are also more expensive than successful ones \citep{fan2025swe}.
Listed token prices can also misrank realized costs because reasoning volume and interaction length differ across models \citep{chen2026price}.
Because reasoning tokens are hidden yet billed, pay-per-token pricing creates a moral hazard that users cannot audit \citep{velasco2025your,sun2025coin}.

Prior work either reports aggregate accuracy across effort levels \citep{kapoor2026holistic}, compares models at a fixed effort \citep{chen2026price,bai2026ai}, or studies a single model on a single-call benchmark \citep{moon2026price}. 
We audit the marginal value of reasoning effort at the task level.
Our contribution brings these strands together in a task-level audit of the marginal value of reasoning effort. 
We analyze agent trajectories across effort levels to uncover how reasoning effort reshapes problem-solving strategies, linking these behavioral changes to task outcomes, inference costs, and modeled waiting costs.
This sharpens the credence-good interpretation of reasoning: even accurately billed computation may be unnecessary, and observing a successful outcome alone does not establish that the additional effort was worth purchasing.

\section{Conclusion}
\label{sec:conclusion}

Reasoning effort is presented to users as an ordered dial, but it does not behave as one.
Across four benchmarks and eight models, we find that higher effort often raises aggregate success yet can still lose tasks that lower effort solves, and on simple tasks solved at every level, higher effort yields much higher inference cost.
Disabling reasoning offers no reliable savings, because agents compensate with longer trajectories that repeatedly reprocess a growing context.
In agentic settings, higher effort can bring fundamental change to the solution strategy rather than merely extending deliberation, often leading to more self-verifications that can sometimes be redundant.

For practitioners, these findings suggest that a fixed effort level is not a cost-effective approach.
A higher effort level should be justified by evidence on the target task rather than by the intuition that more thinking helps, and disabling reasoning should not be assumed to save money in agentic settings.
Where a reliable correctness check exists, na\"ive laddering, which starts at the lowest effort and escalates only after failure, matches or improves on fixed top effort for every model and benchmark we evaluate, at lower median cost on AIME, GPQA, and corrected SWE-Bench Pro.
Practitioners should also evaluate configurations by their full cost, including user waiting time, because token charges alone can misrank them.

More broadly, reasoning effort remains a credence good: users observe neither how much reasoning they receive nor whether it contributed to the outcome.
Providers should make reasoning charges verifiable, guiding customers' decisions, and specify what each effort level controls, in token budget, trajectory length, and behavior, so that effort becomes a controlled quantity rather than a named tier.
The gap between simple effort routing solutions and the oracle router shows that selecting the optimal effort level for each task remains an open problem, since models cannot yet select effort from the task description and laddering depends on external feedback that many deployments lack.
To close this gap, future work can focus on methods that adapt reasoning effort dynamically during execution, spending additional computation only when the difficulty an agent encounters warrants it.

\section*{Acknowledgments}
We thank our colleagues at Together AI and James Zou for their helpful feedback.

\bibliography{custom}
\bibliographystyle{iclr2027_conference}

\clearpage
\appendix

\section{Benchmark details}\label{sec:appendix-benchmarks}

For SWE-Bench-Pro, we sample 44 instances by deterministically selecting four evenly spaced entries from the lexicographically sorted list of instance IDs for each of the benchmark's 11 repositories.

\begin{table}[t]
\centering
\caption{Benchmarks and execution settings.
AIME uses ten attempts per problem and GPQA uses four attempts per question; the agentic benchmarks evaluate multi-turn executions with fixed harnesses.}
\label{tab:benchmarks}
\begin{tabular}{@{}p{0.34\textwidth} p{0.20\textwidth} p{0.38\textwidth}@{}}
\toprule
\textbf{Benchmark} & \textbf{Evaluation size} & \textbf{Execution setting} \\
\midrule
AIME 2026 \cite{dekoninck2026matharena} & 30 problems, 10 attempts each & Single API call per attempt \\
\addlinespace
GPQA Diamond \cite{rein2023gpqa} & 198 questions, 4 attempts each & Single API call per attempt \\
\addlinespace
Terminal-Bench 2.1 \cite{merrill2026terminal} & 89 tasks & Multi-turn terminal agent using Terminus-2 \\
\addlinespace
SWE-Bench Pro \cite{deng2025swe} & 44-instance pilot & Multi-turn repository-level agent using mini-swe-agent \\
\bottomrule
\end{tabular}
\end{table}

\section{Inference setup}\label{sec:appendix-inference}

Table~\ref{tab:inference} lists the API through which each model is served and the reasoning-effort settings we evaluate.
The available settings vary across models: for example, GLM-5.2 supports only \texttt{high} and \texttt{max}, while Kimi K3 supports only reasoning-enabled modes.
For the proprietary models, we evaluate all settings except for \texttt{max} because the extremely high price makes it impractical to evaluate.
Kimi K3 is evaluated on all four benchmarks at \texttt{low}, \texttt{high}, and \texttt{max}.

\begin{table}[t]
\centering
\small
\caption{Serving API and evaluated reasoning-effort settings for each model.}
\label{tab:inference}
\begin{tabular}{@{}lll@{}}
\toprule
\textbf{Model} & \textbf{API} & \textbf{Effort settings} \\
\midrule
GPT-5.6-sol, GPT-5.6-terra, GPT-5.6-luna & OpenAI & \texttt{none}, \texttt{low}, \texttt{medium}, \texttt{high}, \texttt{xhigh} \\
Claude Sonnet 5 & Anthropic & \texttt{none}, \texttt{low}, \texttt{medium}, \texttt{high}, \texttt{xhigh} \\
GLM-5.2 & Together AI & \texttt{none}, \texttt{high}, \texttt{max} \\
Inkling & Together AI & \texttt{none}, \texttt{low}, \texttt{medium}, \texttt{high}, \texttt{max} \\
Kimi K3 & Together AI & \texttt{low}, \texttt{high}, \texttt{max} \\
DeepSeek V4 Flash & Together AI & \texttt{none}, \texttt{low}, \texttt{high}, \texttt{max} \\
\bottomrule
\end{tabular}
\end{table}

\paragraph{Effort control.}
For the GPT-5.6 models, we specify the selected setting via \texttt{reasoning.effort} in the Responses API for AIME and GPQA, and via \texttt{reasoning\_effort} through LiteLLM for the agentic benchmarks.
For Claude Sonnet 5, \texttt{none} explicitly disables extended thinking; all other settings enable adaptive thinking, with \texttt{output\_config.effort} set to the selected level.
For models hosted on Together, we include \texttt{reasoning\_effort} in the request body.

\paragraph{Sampling and output limits.}
We use the providers' default decoding parameters, except for the Terminus-2 agent, for which we retain its default temperature of 0.7.
On AIME and GPQA, we set an output limit of 128{,}000 tokens per call for all models and allow up to five attempts for calls that encounter API errors, using exponential backoff.
On SWE-Bench Pro, we set an output limit of 32{,}000 tokens per call and a spending cap of \$5 per agent execution.
For Terminal-Bench, we retain the Terminus-2 limits except for agent timeout multipliers of $3\times$ for GLM-5.2 and Inkling, $5\times$ for Kimi K3, and $9\times$ for DeepSeek V4 Flash, setup based on the availability of TogetherAPI call. 
These are held fixed across each model's effort settings.

\paragraph{Single-call benchmarks.}
For each AIME attempt, we send the MathArena instruction and the problem in a single user message, without a system prompt, and extract the answer from the final \verb|\boxed{}| expression in the response.
For each GPQA attempt, we use the multiple-choice prompt from OpenAI's \texttt{simple-evals}\footnote{\url{https://github.com/openai/simple-evals}}, shuffle the answer choices independently, and extract the answer from the \texttt{Answer:} line.

\paragraph{Agentic benchmarks.}
We evaluate Terminal-Bench 2.1 using the \texttt{harbor} harness and the Terminus-2 agent under the official configuration, with one trial per task and no turn limit beyond the task-specific wall-clock timeout.
We evaluate SWE-Bench Pro using \texttt{mini-swe-agent} under the benchmark's official configuration, with one run per instance, a maximum of 250 model calls per instance, and a 60-second timeout per shell command.
The GPT-5.6 models, Claude Sonnet 5, and GLM-5.2 use the text-based scaffold, which requires exactly one bash code block per response.

\section{Experimental cost}\label{sec:appendix-cost}

Table~\ref{tab:app-cost} includes an approximation of the cost of this study, including developmental cost and finalized experiments.
To keep the comparison across models and reasoning-effort settings within budget, we restrict the SWE-Bench Pro evaluation to 44 instances spanning all 11 repositories.

\begin{table}[t]
\centering
\small
\setlength{\tabcolsep}{4pt}
\caption{Experimental cost in USD, rounded to one decimal place.}
\label{tab:app-cost}
\begin{tabular}{@{}lrrrrr@{}}
\toprule
\textbf{Model / cost category} & \textbf{AIME} & \textbf{GPQA} & \textbf{SWE-Pro} & \textbf{TB} & \textbf{Total (USD)} \\
\midrule
\multicolumn{6}{@{}l}{\textbf{Benchmark inference}} \\
\addlinespace[2pt]
GPT-5.6-sol & 72.4 & 125.9 & 1{,}142.7 & 165.1 & 1{,}506.0 \\
GPT-5.6-terra & 29.2 & 47.6 & 312.1 & 131.8 & 520.8 \\
GPT-5.6-luna & 4.9 & 10.2 & 15.4 & 19.3 & 49.8 \\
Claude Sonnet 5 & 217.2 & 230.3 & 1{,}045.2 & 322.3 & 1{,}815.0 \\
GLM-5.2 & 44.7 & 126.7 & 804.9 & 255.7 & 1{,}232.1 \\
Inkling & 60.0 & 104.8 & 416.3 & 412.3 & 993.4 \\
Kimi K3 & 128.5 & 150.4 & 236.2 & 216.2 & 731.4 \\
DeepSeek V4 Flash & 3.0 & 4.4 & 107.9 & 510.3 & 625.6 \\
\addlinespace[2pt]
\textbf{Inference subtotal} & \textbf{559.9} & \textbf{800.2} & \textbf{4{,}080.8} & \textbf{2{,}033.1} & \textbf{7{,}474.0} \\
\midrule
\multicolumn{5}{@{}l}{Scaffold sensitivity tests} & 218.7 \\
\multicolumn{5}{@{}l}{\actonomy{} annotation (2{,}067 retained annotations)} & 3{,}857.6 \\
\multicolumn{5}{@{}l}{Self-routing (12{,}026 retained selection calls)} & 44.2 \\
\midrule
\multicolumn{5}{@{}l}{Developmental cost: inference and routing} & 2{,}012.3 \\
\multicolumn{5}{@{}l}{Developmental cost: annotation} & $\approx$464.2 \\
\multicolumn{5}{@{}l}{\textbf{Developmental subtotal}} & $\approx$\textbf{2{,}476.5} \\
\midrule
\multicolumn{5}{@{}l}{\textbf{Total API-equivalent cost}} & $\approx$\textbf{14{,}071.1} \\
\bottomrule
\end{tabular}
\end{table}

\section{Overall performance across effort levels}\label{sec:appendix-performance}

Table~\ref{tab:app_perf} reports overall accuracy for each model and effort setting, including both the original and corrected SWE-Bench Pro pilots.
Terminal-Bench trials without a successful verifier verdict are counted as failures, and provider refusals on GPQA are counted as incorrect attempts.
For SWE-Bench Pro, we follow the benchmark's scoring rule: executions that pass the evaluator but cost more than \$5 are counted as failures.

\begin{table}[p]
    \centering
    \small
    \setlength{\tabcolsep}{3pt}
    \caption{Overall accuracy (\%) across reasoning-effort settings, with 95\% confidence intervals in brackets, rounded to whole percentage points.
    AIME and GPQA report mean per-attempt correctness, with percentile bootstrap intervals (10,000 resamples) over problems that keep each problem's attempts together; the agentic benchmarks report the fraction of tasks solved, with Wilson score intervals.
    Intervals describe each setting separately; because all settings of a model share the same tasks, paired tests between settings are more sensitive than a comparison of overlapping intervals.
    Bold marks the highest accuracy within each model--benchmark row, including ties; dashes indicate settings without a reported result.
    Effort names retain their model-specific meanings.}
    \label{tab:app_perf}
    \begin{tabular}{@{}lrrrrrr@{}}
        \toprule
        Model & \texttt{none} & \texttt{low} & \texttt{medium} & \texttt{high} & \texttt{xhigh} & \texttt{max} \\
        \midrule
        \multicolumn{7}{@{}l}{\textbf{AIME 2026 (30 problems; 10 attempts per problem)}} \\
        \addlinespace[2pt]
        GPT-5.6-luna & 25.3\,{\scriptsize[14, 38]} & 75.7\,{\scriptsize[65, 85]} & 85.3\,{\scriptsize[75, 94]} & 94.7\,{\scriptsize[89, 99]} & \textbf{96.7}\,{\scriptsize[93, 99]} & -- \\
        GPT-5.6-terra & 7.3\,{\scriptsize[2, 14]} & 89.7\,{\scriptsize[81, 96]} & 91.3\,{\scriptsize[82, 98]} & 92.3\,{\scriptsize[86, 98]} & \textbf{97.0}\,{\scriptsize[94, 99]} & -- \\
        GPT-5.6-sol & 55.3\,{\scriptsize[42, 68]} & 96.7\,{\scriptsize[90, 100]} & 97.7\,{\scriptsize[95, 100]} & \textbf{99.7}\,{\scriptsize[99, 100]} & \textbf{99.7}\,{\scriptsize[99, 100]} & -- \\
        Claude Sonnet 5 & 79.7\,{\scriptsize[68, 90]} & 72.3\,{\scriptsize[60, 83]} & 86.3\,{\scriptsize[78, 93]} & 96.7\,{\scriptsize[92, 99]} & \textbf{97.3}\,{\scriptsize[92, 100]} & -- \\
        GLM-5.2 & 46.0\,{\scriptsize[35, 57]} & -- & -- & 86.3\,{\scriptsize[76, 95]} & -- & \textbf{90.7}\,{\scriptsize[81, 98]} \\
        Inkling & 25.7\,{\scriptsize[18, 34]} & 77.7\,{\scriptsize[67, 87]} & 90.7\,{\scriptsize[81, 98]} & 93.7\,{\scriptsize[86, 99]} & -- & \textbf{94.3}\,{\scriptsize[86, 100]} \\
        Kimi K3 & -- & 65.7\,{\scriptsize[54, 76]} & -- & 89.0\,{\scriptsize[84, 93]} & -- & \textbf{90.7}\,{\scriptsize[85, 95]} \\
        DeepSeek V4 Flash & 65.0\,{\scriptsize[52, 77]} & 96.3\,{\scriptsize[90, 100]} & -- & \textbf{97.3}\,{\scriptsize[95, 99]} & -- & 95.3\,{\scriptsize[92, 98]} \\
        \midrule
        \multicolumn{7}{@{}l}{\textbf{GPQA Diamond (198 questions; 4 attempts per question)}} \\
        \addlinespace[2pt]
        GPT-5.6-luna & 64.0\,{\scriptsize[58, 70]} & 81.8\,{\scriptsize[77, 86]} & 84.8\,{\scriptsize[81, 89]} & 88.9\,{\scriptsize[85, 92]} & \textbf{90.0}\,{\scriptsize[86, 94]} & -- \\
        GPT-5.6-terra & 72.6\,{\scriptsize[67, 78]} & 86.1\,{\scriptsize[82, 90]} & 87.9\,{\scriptsize[84, 91]} & 89.0\,{\scriptsize[85, 92]} & \textbf{89.8}\,{\scriptsize[86, 93]} & -- \\
        GPT-5.6-sol & 82.4\,{\scriptsize[78, 87]} & 91.0\,{\scriptsize[87, 94]} & 90.7\,{\scriptsize[87, 94]} & 92.3\,{\scriptsize[89, 95]} & \textbf{93.1}\,{\scriptsize[90, 96]} & -- \\
        Claude Sonnet 5 & 81.8\,{\scriptsize[77, 86]} & 78.8\,{\scriptsize[74, 83]} & 84.6\,{\scriptsize[80, 89]} & 88.8\,{\scriptsize[85, 92]} & \textbf{90.0}\,{\scriptsize[86, 94]} & -- \\
        GLM-5.2 & 66.7\,{\scriptsize[62, 71]} & -- & -- & 89.4\,{\scriptsize[86, 93]} & -- & \textbf{91.9}\,{\scriptsize[89, 95]} \\
        Inkling & 67.4\,{\scriptsize[62, 73]} & 65.3\,{\scriptsize[60, 71]} & 81.2\,{\scriptsize[77, 85]} & 84.3\,{\scriptsize[80, 88]} & -- & \textbf{87.9}\,{\scriptsize[84, 91]} \\
        Kimi K3 & -- & 84.2\,{\scriptsize[80, 88]} & -- & 91.5\,{\scriptsize[88, 95]} & -- & \textbf{94.3}\,{\scriptsize[91, 97]} \\
        DeepSeek V4 Flash & 73.4\,{\scriptsize[68, 78]} & 89.8\,{\scriptsize[86, 93]} & -- & 89.9\,{\scriptsize[86, 93]} & -- & \textbf{91.4}\,{\scriptsize[88, 95]} \\
        \midrule
        \multicolumn{7}{@{}l}{\textbf{Terminal-Bench 2.1 (89 tasks)}} \\
        \addlinespace[2pt]
        GPT-5.6-luna & 39.3\,{\scriptsize[30, 50]} & 49.4\,{\scriptsize[39, 60]} & 52.8\,{\scriptsize[43, 63]} & 62.9\,{\scriptsize[53, 72]} & \textbf{70.8}\,{\scriptsize[61, 79]} & -- \\
        GPT-5.6-terra & 60.7\,{\scriptsize[50, 70]} & 65.2\,{\scriptsize[55, 74]} & 68.5\,{\scriptsize[58, 77]} & \textbf{74.2}\,{\scriptsize[64, 82]} & \textbf{74.2}\,{\scriptsize[64, 82]} & -- \\
        GPT-5.6-sol & 66.3\,{\scriptsize[56, 75]} & 76.4\,{\scriptsize[67, 84]} & 76.4\,{\scriptsize[67, 84]} & \textbf{85.4}\,{\scriptsize[77, 91]} & 82.0\,{\scriptsize[73, 89]} & -- \\
        Claude Sonnet 5 & \textbf{75.3}\,{\scriptsize[65, 83]} & 57.3\,{\scriptsize[47, 67]} & 64.0\,{\scriptsize[54, 73]} & 70.8\,{\scriptsize[61, 79]} & \textbf{75.3}\,{\scriptsize[65, 83]} & -- \\
        GLM-5.2 & 58.4\,{\scriptsize[48, 68]} & -- & -- & \textbf{71.9}\,{\scriptsize[62, 80]} & -- & \textbf{71.9}\,{\scriptsize[62, 80]} \\
        Inkling & \textbf{42.7}\,{\scriptsize[33, 53]} & 31.5\,{\scriptsize[23, 42]} & 29.2\,{\scriptsize[21, 39]} & 36.0\,{\scriptsize[27, 46]} & -- & 37.1\,{\scriptsize[28, 47]} \\
        Kimi K3 & -- & 77.5\,{\scriptsize[68, 85]} & -- & \textbf{80.9}\,{\scriptsize[72, 88]} & -- & 73.0\,{\scriptsize[63, 81]} \\
        DeepSeek V4 Flash & 65.2\,{\scriptsize[55, 74]} & 56.2\,{\scriptsize[46, 66]} & -- & 64.0\,{\scriptsize[54, 73]} & -- & \textbf{69.7}\,{\scriptsize[59, 78]} \\
        \midrule
        \multicolumn{7}{@{}l}{\textbf{SWE-Bench Pro, original (44 instances)}} \\
        \addlinespace[2pt]
        GPT-5.6-luna & 52.3\,{\scriptsize[38, 66]} & 38.6\,{\scriptsize[26, 53]} & 54.5\,{\scriptsize[40, 68]} & 65.9\,{\scriptsize[51, 78]} & \textbf{84.1}\,{\scriptsize[71, 92]} & -- \\
        GPT-5.6-terra & 45.5\,{\scriptsize[32, 60]} & 56.8\,{\scriptsize[42, 70]} & 63.6\,{\scriptsize[49, 76]} & 65.9\,{\scriptsize[51, 78]} & \textbf{72.7}\,{\scriptsize[58, 84]} & -- \\
        GPT-5.6-sol & 65.9\,{\scriptsize[51, 78]} & 61.4\,{\scriptsize[47, 74]} & 72.7\,{\scriptsize[58, 84]} & 72.7\,{\scriptsize[58, 84]} & \textbf{77.3}\,{\scriptsize[63, 87]} & -- \\
        Claude Sonnet 5 & 59.1\,{\scriptsize[44, 72]} & 54.5\,{\scriptsize[40, 68]} & 52.3\,{\scriptsize[38, 66]} & 61.4\,{\scriptsize[47, 74]} & \textbf{79.5}\,{\scriptsize[65, 89]} & -- \\
        GLM-5.2 & \textbf{72.7}\,{\scriptsize[58, 84]} & -- & -- & 65.9\,{\scriptsize[51, 78]} & -- & 70.5\,{\scriptsize[56, 82]} \\
        Inkling & 63.6\,{\scriptsize[49, 76]} & 18.2\,{\scriptsize[10, 32]} & 52.3\,{\scriptsize[38, 66]} & 61.4\,{\scriptsize[47, 74]} & -- & \textbf{65.9}\,{\scriptsize[51, 78]} \\
        Kimi K3 & -- & 65.9\,{\scriptsize[51, 78]} & -- & 79.5\,{\scriptsize[65, 89]} & -- & \textbf{93.2}\,{\scriptsize[82, 98]} \\
        DeepSeek V4 Flash & 52.3\,{\scriptsize[38, 66]} & 59.1\,{\scriptsize[44, 72]} & -- & \textbf{84.1}\,{\scriptsize[71, 92]} & -- & \textbf{84.1}\,{\scriptsize[71, 92]} \\
        \midrule
        \multicolumn{7}{@{}l}{\textbf{SWE-Bench Pro, corrected (44 instances)}} \\
        \addlinespace[2pt]
        GPT-5.6-luna & 40.9\,{\scriptsize[28, 56]} & 47.7\,{\scriptsize[34, 62]} & 47.7\,{\scriptsize[34, 62]} & 52.3\,{\scriptsize[38, 66]} & \textbf{56.8}\,{\scriptsize[42, 70]} & -- \\
        GPT-5.6-terra & 52.3\,{\scriptsize[38, 66]} & 50.0\,{\scriptsize[36, 64]} & 52.3\,{\scriptsize[38, 66]} & 50.0\,{\scriptsize[36, 64]} & \textbf{61.4}\,{\scriptsize[47, 74]} & -- \\
        GPT-5.6-sol & 43.2\,{\scriptsize[30, 58]} & 59.1\,{\scriptsize[44, 72]} & 56.8\,{\scriptsize[42, 70]} & \textbf{61.4}\,{\scriptsize[47, 74]} & 59.1\,{\scriptsize[44, 72]} & -- \\
        Claude Sonnet 5 & 50.0\,{\scriptsize[36, 64]} & \textbf{56.8}\,{\scriptsize[42, 70]} & 52.3\,{\scriptsize[38, 66]} & 47.7\,{\scriptsize[34, 62]} & 50.0\,{\scriptsize[36, 64]} & -- \\
        GLM-5.2 & 50.0\,{\scriptsize[36, 64]} & -- & -- & \textbf{54.5}\,{\scriptsize[40, 68]} & -- & 43.2\,{\scriptsize[30, 58]} \\
        Inkling & 43.2\,{\scriptsize[30, 58]} & 25.0\,{\scriptsize[15, 39]} & 38.6\,{\scriptsize[26, 53]} & 45.5\,{\scriptsize[32, 60]} & -- & \textbf{54.5}\,{\scriptsize[40, 68]} \\
        Kimi K3 & -- & 54.5\,{\scriptsize[40, 68]} & -- & 52.3\,{\scriptsize[38, 66]} & -- & \textbf{70.5}\,{\scriptsize[56, 82]} \\
        DeepSeek V4 Flash & 56.8\,{\scriptsize[42, 70]} & 54.5\,{\scriptsize[40, 68]} & -- & 61.4\,{\scriptsize[47, 74]} & -- & \textbf{72.7}\,{\scriptsize[58, 84]} \\
        \bottomrule
    \end{tabular}
\end{table}

\section{Instances gained and lost on the remaining datasets}\label{sec:appendix-instances-gained}

Figure~\ref{fig:app_instances_gained} presents the comparison shown in Figure~\ref{fig:instances_gained} for AIME, GPQA, and the corrected SWE-Bench Pro environments.
For each pair of adjacent effort levels, solid bars indicate the number of instances solved at the higher effort level but not at the lower level; hollow bars indicate the number solved at the lower level but not at the higher level.

\begin{figure}[t]
    \centering
    \includegraphics[width=\linewidth]{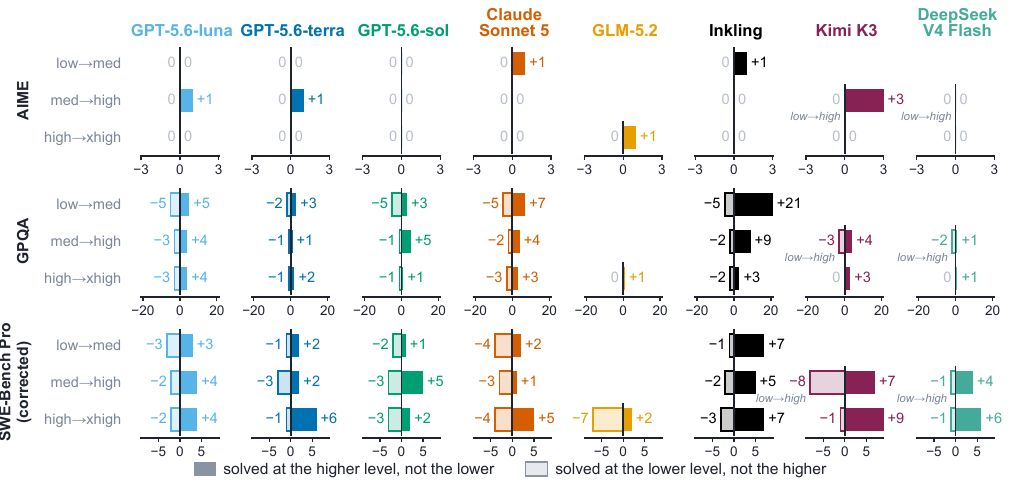}
    \caption{
    Instances gained and lost between adjacent reasoning-effort settings on AIME, GPQA, and the corrected SWE-Bench Pro environments.
    An instance counts as solved at a level if any of its attempts at that level is correct.}
    \label{fig:app_instances_gained}
\end{figure}

\section{User waiting cost dominates the bill}\label{sec:appendix-full-cost}

The inference costs reported in \S\ref{sec:cost} exclude the opportunity cost of the time users spend waiting for an agent to complete its execution.
Figure~\ref{fig:solvable_cost_full} extends the always-solved cost comparison in Figure~\ref{fig:solvable_cost} by adding this waiting cost to the API charges, following the full-cost formulation in Equation~\ref{eq:full-cost}.
There, $C_{\mathrm{API}}$ is the inference bill in U.S. dollars, $N_{\mathrm{out}}$ counts all generated tokens across the execution, including reasoning tokens, and $r_m$ is the model's tabulated output throughput in tokens per second.
We set the hourly wage $w = 33.54$, the average wage across all U.S. occupations.\footnote{%
Source: U.S. Bureau of Labor Statistics, Occupational Employment and Wage Statistics, \href{https://www.bls.gov/news.release/archives/ocwage_05152026.pdf}{May 2025 national estimates, Table 1}, which reports a mean hourly wage of \$33.54 across all occupations. While there are many caveats to this analysis, we believe it is an approximate estimate of the time cost required to solve problem instances.}
The factor of $3600$ converts decoding time from seconds to hours.
This estimate assumes that the user is blocked throughout the generation and accounts only for decoding time; it excludes prompt processing, queueing, and tool-execution latency. 
If these additional delays also block users, the corresponding waiting cost would be higher.

Under these assumptions, \textbf{waiting cost accounts for approximately 58--98\% of the combined cost} across the agentic model--effort comparisons in the figure.
For example, at the highest effort level, GPT-5.6-sol incurs mean API charges of \$1.24 on the always-solved SWE-Bench Pro set; including waiting raises the mean cost to \$4.62, of which 73\% is attributable to user time.
Thus, even when valued at the mean wage across all occupations, the modeled cost of user time can exceed the API charges.
In practice, users may perform other activities while an agent is running \citep{zheng2024fragmented}.
However, idle waiting persists in the use of coding agents \citep{robbes2026agentic}, and context switching can impose additional cognitive costs, including increased stress and impaired subsequent performance \citep{mark2008cost,leroy2009so}.
These effects are not captured by our estimate.

Figure~\ref{fig:app_self_routing_full} extends the routing comparisons in Figures~\ref{fig:self_routing} and~\ref{fig:app_self_routing} by replacing API charges with full cost.
Relative to fixed top effort, including waiting cost increases laddering's savings from 17.4\% to 26.3\% for GPT-5.6-sol, from 4.8\% to 19.5\% for GPT-5.6-terra, and from 8.7\% to 18.1\% for Claude Sonnet 5.
For GPT-5.6-luna, an 8.2\% cost increase becomes a 5.1\% saving.
Laddering remains more costly for GLM-5.2, Inkling, and DeepSeek V4 Flash, but the respective cost increases decline from 38.8\%, 11.5\%, and 88.3\% to 20.1\%, 7.3\%, and 54.2\%.
Kimi K3 remains cheaper under laddering, although including waiting reduces its saving from 30.5\% to 19.6\%.
Using the mean hourly wage for software developers (\$71.20) yields a median saving of 11.5\% across the eight models, compared with 11.6\% at the mean wage across all occupations.

\begin{figure}[t]
    \centering
    \includegraphics[width=\linewidth]{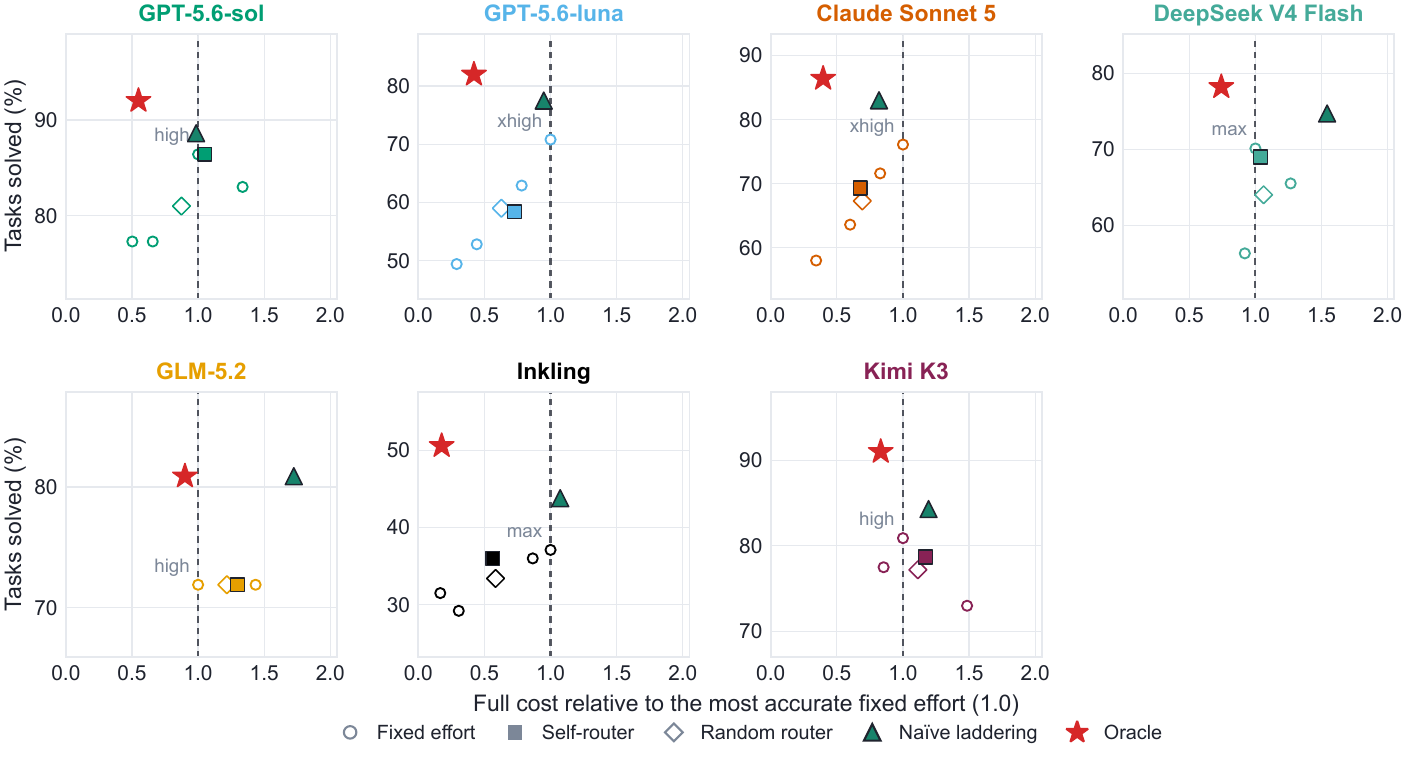}
    \caption{Self-routing and na\"ive laddering on Terminal-Bench 2.1 with user waiting cost added to the API bill (Equation~\ref{eq:full-cost}), for the seven models with saved self-routing comparisons.
    Each panel shows success against full cost, normalized by the full cost of that model's fixed effort with the highest performance (dashed line).
    The oracle selects the successful effort with the lowest full cost; router calls are charged their API cost only.
    Policies, markers, and task cohorts match Figure~\ref{fig:self_routing}.
    Lower cost and higher success are preferable.}
    \label{fig:app_self_routing_full}
\end{figure}

\begin{figure}[t]
    \centering
    \includegraphics[width=\linewidth]{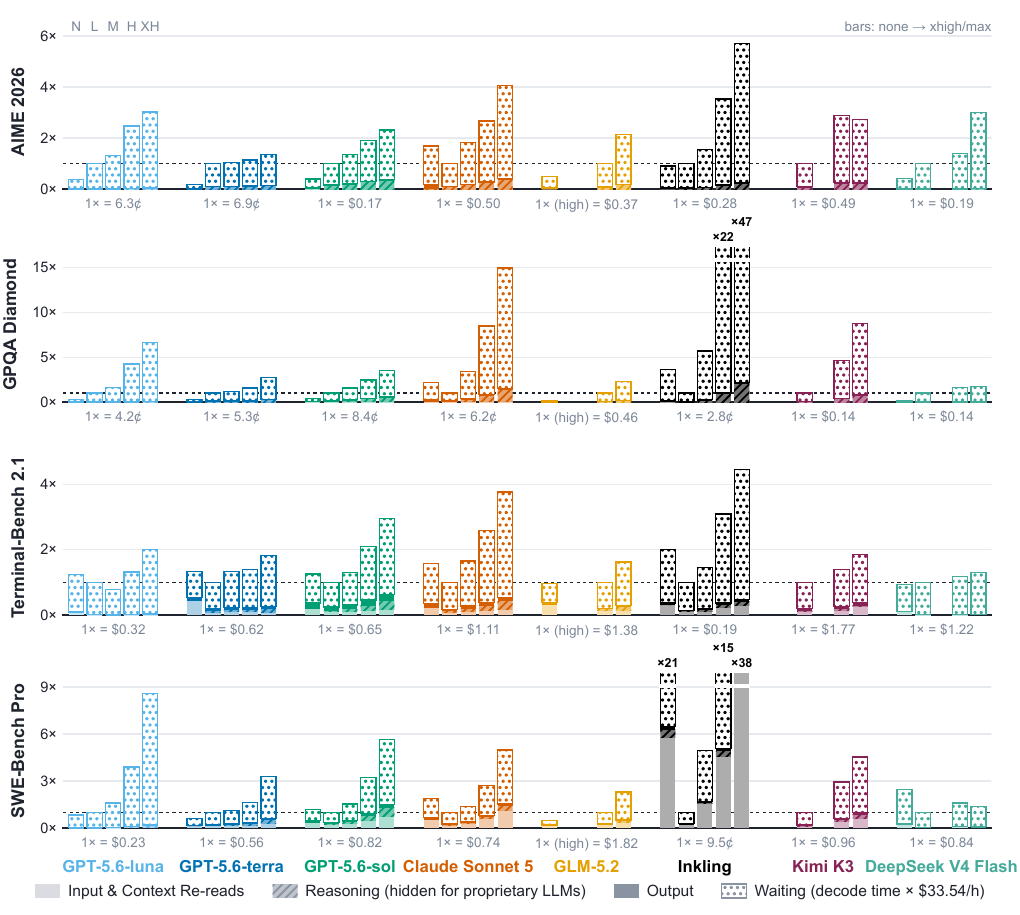}
    \caption{Full cost of one task the model solves at every effort level, including \texttt{none} if supported, relative to the full cost at \texttt{low} (dashed line; GLM-5.2 is anchored on \texttt{high}).
    Kimi K3 has no \texttt{none} setting.
    Full cost adds to the API bill the time a user waits for the output tokens at the model's published throughput, priced at the BLS mean hourly wage of \$33.54.
    Each bar is split into input, hidden reasoning, visible output, and waiting.
    The always-solved set is fixed per model and benchmark; the footer gives the dollar value of $1\times$.
    Bars taller than the axis are clipped and labelled.}
    \label{fig:solvable_cost_full}
\end{figure}

\section{Corrected SWE-Bench Pro environment}\label{sec:appendix-swebenchpro-example}

Table~\ref{tab:swebenchpro-corrected-perf} compares performance in the original and corrected environments at each model's highest evaluated reasoning-effort setting.
All eight models achieve lower solve rates in the corrected environment, with declines of 11.4--29.5 percentage points.
Because the two environments use the same instances, we assess each decline using an exact McNemar test on discordant outcomes, i.e., instances solved in one environment but not the other.
The declines are significant for five of the eight models at $\alpha = 0.05$ and remain significant after Holm correction.
A sign test across the eight models yields $p = 0.008$.
These results indicate that the environment corrections substantially affected the agents' performance, raising questions about the validity of evaluation results using the official SWE-Bench-Pro release.

\begin{table}[t]
    \centering
    \small
    \setlength{\tabcolsep}{5pt}
    \caption{Solve rates on the original and corrected SWE-Bench Pro environments at each model's highest evaluated reasoning-effort setting, using the same 44 instances in both environments.
    Corrected removes the task repository's original git history and restricts network access to allowlisted package registries.
    $\Delta$ is corrected minus original solve rate in percentage points, computed before rounding.
    The \emph{Solved only in} columns count discordant instances, and $p$ is the two-sided exact McNemar test on those counts.
    The decrease is significant at $\alpha = 0.05$ for 5 of 8 models, and a sign test over the 8 models, all of which decrease, gives $p = 0.008$.
    Both conditions follow the source scoring rule that evaluator-passed executions costing more than \$5 count as failures.
    Table~\ref{tab:app_perf} reports results at all effort levels.}
    \label{tab:swebenchpro-corrected-perf}
    \begin{tabular}{@{}llrrrrrr@{}}
        \toprule
        & & & & & \multicolumn{2}{c}{Solved only in} & \\
        \cmidrule(lr){6-7}
        Model & Effort & Original (\%) & Corrected (\%) & $\Delta$ (pp) & Original & Corrected & $p$ \\
        \midrule
        GPT-5.6-luna & \texttt{xhigh} & 84.1 & 56.8 & $-27.3$ & 13 & 1 & 0.002 \\
        GPT-5.6-terra & \texttt{xhigh} & 72.7 & 61.4 & $-11.4$ & 7 & 2 & 0.180 \\
        GPT-5.6-sol & \texttt{xhigh} & 77.3 & 59.1 & $-18.2$ & 8 & 0 & 0.008 \\
        Claude Sonnet 5 & \texttt{xhigh} & 79.5 & 50.0 & $-29.5$ & 13 & 0 & $<$0.001 \\
        GLM-5.2 & \texttt{max} & 70.5 & 43.2 & $-27.3$ & 13 & 1 & 0.002 \\
        Inkling & \texttt{max} & 65.9 & 54.5 & $-11.4$ & 5 & 0 & 0.062 \\
        Kimi K3 & \texttt{max} & 93.2 & 70.5 & $-22.7$ & 11 & 1 & 0.006 \\
        DeepSeek V4 Flash & \texttt{max} & 84.1 & 72.7 & $-11.4$ & 6 & 1 & 0.125 \\
        \bottomrule
    \end{tabular}
\end{table}

Figure~\ref{fig:app_swebenchpro_example} illustrates an instance in which the original SWE-Bench Pro container retains the gold fix commit in its Git history.
At the highest effort level, the agent locates this commit and copies its patch; at the lowest effort level, it solves the same task without using the commit.
Both runs are scored as resolved, yet the higher-effort run costs $5.3\times$ as much and reproduces an existing fix rather than independently deriving one.

\begin{figure}[t]
    \centering
    \includegraphics[width=\linewidth]{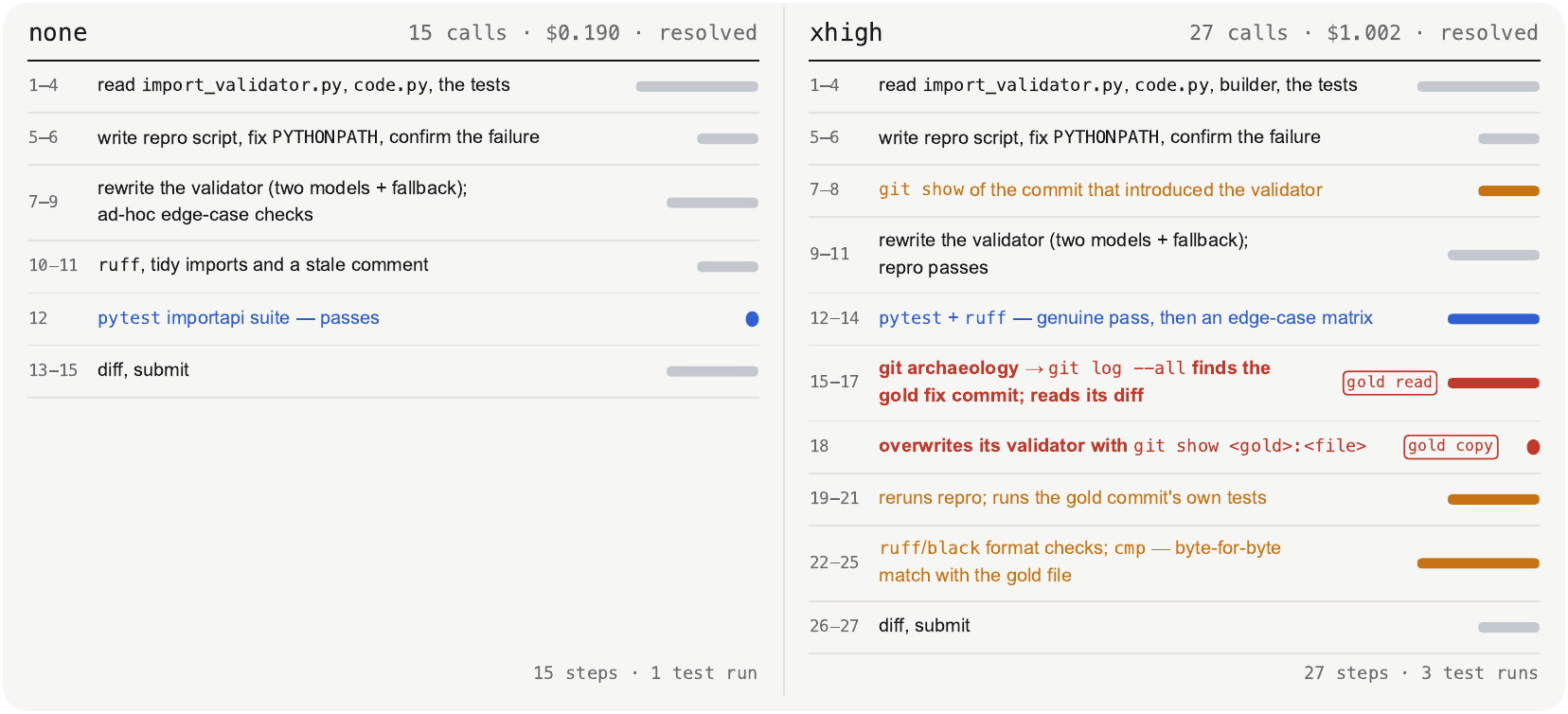}
    \caption{The summarized trajectory of GPT-5.6-sol on the SWE-Bench Pro instance \texttt{openlibrary-00bec1e7} at \texttt{none} (left, 15 calls, \$0.19) and \texttt{xhigh} (right, 27 calls, \$1.00); both runs resolve the task. 
    Each row is a phase of consecutive agent steps and bar length is the number of steps in that phase. 
    Grey marks routine work, blue a genuine test run, orange a look into git history or validation against the copied file, and red a read or copy of the gold fix commit. At \texttt{none} the agent fixes the validator from the issue text and the code. At \texttt{xhigh} it does the same work, then runs \texttt{git log --all}, finds the gold fix commit still present in the container's clone, overwrites its own file with \texttt{git show <gold>:<file>}, runs that commit's tests, and confirms a byte-for-byte match before submitting.}
    \label{fig:app_swebenchpro_example}
\end{figure}

Figure~\ref{fig:app-swebenchpro-corrected-cost} repeats the always-solved API-cost comparison of Figure~\ref{fig:solvable_cost} in the corrected SWEBench-Pro environment.
The always-solved sets are computed separately within this environment and contain 10--20 of the 44 instances per model.
For all eight models, the highest evaluated effort setting costs more than the lowest reasoning setting on these fixed task sets, with ratios ranging from $1.83\times$ to $25.59\times$.
The increase is again not uniformly monotonic: DeepSeek V4 Flash costs less at \texttt{max} than at \texttt{high}, while both remain more costly than \texttt{low}.

\begin{figure}[t]
    \centering
    \includegraphics[width=\linewidth]{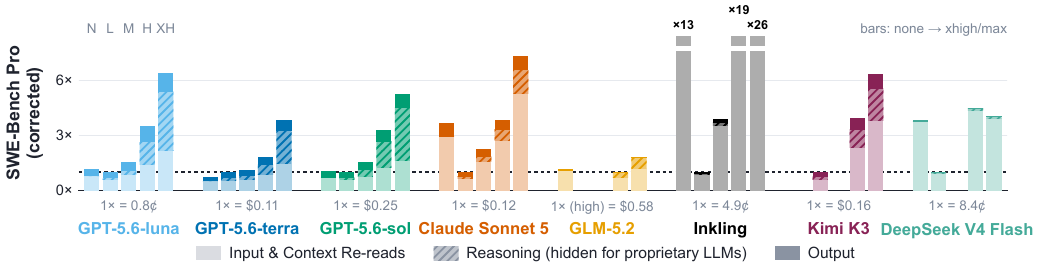}
    \caption{Mean API cost per always-solved task in the corrected SWE-Bench Pro environment, using the same 44-instance sample as the original environment.
    Bars follow the order \texttt{none}, \texttt{low}, \texttt{medium}, \texttt{high}, and \texttt{xhigh}/\texttt{max}, leaving unsupported settings empty, and are split into input, hidden reasoning, and visible output charges.
    Bars taller than the axis are clipped and labelled.}
    \label{fig:app-swebenchpro-corrected-cost}
\end{figure}

\section{Evaluation of na\"ive laddering}\label{sec:appendix-laddering}

We reconstruct laddering success rates and API costs from our existing executions, counting the costs of both attempts whenever escalation occurs.
Each policy execution consists of an initial attempt at the lowest available reasoning-effort setting and, if that attempt fails, at most one fresh retry at the highest setting.
On AIME and GPQA, we pair attempts across effort levels by problem and attempt index.
The ten attempts per AIME problem and four per GPQA question therefore yield repeated policy evaluations; they are not treated as a batch that succeeds whenever any attempt is correct.

Laddering includes the full recorded costs and applies no additional cap to the combined cost of an initial attempt and its retry.
The analysis assumes a reliable external correctness verdict and evaluates a fresh retry, rather than self-detection of failure or continuation of the initial trajectory.
Preserving successful initial attempts makes the recorded success set the union of the starting and highest effort levels' successes.
Cost savings depend on whether avoiding high-effort runs on initial successes offsets the cost of failed first attempts.

\subsection{Ablation: same-effort retries on DeepSWE}\label{sec:appendix-retry-controls}

To verify that laddering's success gains are not simply due to the availability of a second attempt, we compare it with two same-effort retry controls.
We compare laddering ($L\!\rightarrow\!H$) with retrying at low effort ($L\!\rightarrow\!L$) or high effort ($H\!\rightarrow\!H$), using the DeepSWE rollouts \citep{huang2026deepswe} described in Appendix~\ref{sec:appendix-repeated-runs}.
The results match our existing analysis and are listed in Table~\ref{tab:app-deepswe-retries}.

We average over all ordered pairs of distinct rollouts within each task, using same-effort pairs for the controls and cross-effort pairs for laddering, then weight tasks equally.
Success means either attempt passes, and the retry is charged only when the first attempt fails.

Across all five models, laddering achieves higher estimated success than $L\!\rightarrow\!L$ and is substantially cheaper than $H\!\rightarrow\!H$ (Table~\ref{tab:app-deepswe-retries}).
Although $H\!\rightarrow\!H$ remains more successful, laddering offers a less expensive way to improve on repeated low-effort attempts.

\begin{table}[t]
    \centering
    \small
    \setlength{\tabcolsep}{5pt}
    \caption{Retry control study on DeepSWE v1.1, averaged equally over 113 tasks per model.
    Each policy makes one initial attempt and at most one fresh retry after failure.
    Success is reported in percent; cost is mean API dollars per task, including retries only when needed.
    $L/H$ denote \texttt{low}/\texttt{xhigh}, except for GLM-5.2 (\texttt{high}/\texttt{max}).}
    \label{tab:app-deepswe-retries}
    \begin{tabular}{@{}lrrrrrr@{}}
        \toprule
        & \multicolumn{2}{c}{$L\!\rightarrow\!L$} & \multicolumn{2}{c}{$L\!\rightarrow\!H$} & \multicolumn{2}{c}{$H\!\rightarrow\!H$} \\
        \cmidrule(lr){2-3}\cmidrule(lr){4-5}\cmidrule(lr){6-7}
        Model & Success & Cost (\$) & Success & Cost (\$) & Success & Cost (\$) \\
        \midrule
        GPT-5.6-luna & 2.8 & 0.029 & 57.1 & 0.308 & 70.6 & 0.425 \\
        GPT-5.6-terra & 34.7 & 0.608 & 64.6 & 1.692 & 72.0 & 2.379 \\
        GPT-5.6-sol & 59.6 & 1.688 & 76.8 & 3.796 & 79.9 & 6.044 \\
        Claude Sonnet 5 & 43.8 & 3.806 & 59.0 & 10.994 & 64.2 & 17.607 \\
        GLM-5.2 & 52.7 & 4.781 & 57.5 & 5.497 & 61.5 & 6.136 \\
        \bottomrule
    \end{tabular}
\end{table}

\section{Self-routing on the remaining models}\label{sec:appendix-self-routing}

Figure~\ref{fig:app_self_routing} repeats the self-routing comparison of Figure~\ref{fig:self_routing} for GLM-5.2, Inkling, and Kimi K3 on Terminal-Bench 2.1.
On the original 44-task SWE-Bench Pro pilot, Kimi K3 self-routing matches random allocation's 79.5\% success rate at 3.1\% lower cost, including selection calls; these results use the original repositories rather than the corrected environment.

\begin{figure}[t]
    \centering
    \includegraphics[width=\linewidth]{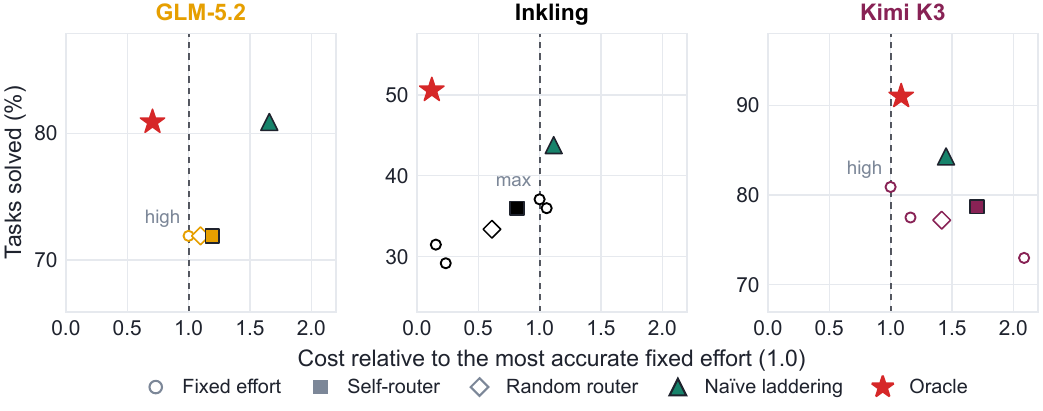}
    \caption{Self-routing and na\"ive laddering compared with fixed efforts, random allocation, and the oracle on Terminal-Bench 2.1 for GLM-5.2, Inkling, and Kimi K3.
    The layout and policies match Figure~\ref{fig:self_routing}.
    Na\"ive laddering starts at \texttt{low} for Inkling and Kimi K3 and at \texttt{high} for GLM-5.2, retrying verified failures at \texttt{max}.
    Lower cost and higher success are preferable.}
    \label{fig:app_self_routing}
\end{figure}

Table~\ref{tab:app_prompt} gives the prompt for the self-router.

\begin{table}[t]
\centering
\small
\newcommand{\ph}[1]{{\normalfont\itshape$\langle$#1$\rangle$}}
\caption{Prompt used for self-routing.
Each selection sends the template in (a) as a single user message, with the fields in angle brackets filled in and the unit name written in capitals in the task delimiters.
The task description is the text the model would receive at the start of the run.
The selection call runs at \texttt{low} effort, or at \texttt{high} for GLM-5.2, which has no \texttt{low} setting.
The chosen level is read from the \texttt{EFFORT:} line, or otherwise from the last option named in the response.}
\label{tab:app_prompt}
\begin{tabular}{@{}p{0.97\textwidth}@{}}
\toprule
\textbf{(a) Prompt template} \\
\midrule
\ttfamily\raggedright
You are about to be given one \ph{unit} to solve on your own.
\par\medskip
Setting: \ph{setting}
\par\medskip
Before you start, choose the reasoning effort you will run at. Your options, from least to most thinking:
\par
\hspace*{1em}\ph{option 1} {\normalfont\textemdash} \ph{description 1}
\par
\hspace*{1em}\ldots
\par
\hspace*{1em}\ph{option $k$} {\normalfont\textemdash} \ph{description $k$}
\par\medskip
Thinking longer costs more money and more time, and on a hard task it is often what makes the difference between solving it and not. Thinking longer on an easy task buys nothing. Choose the LOWEST level you believe will still let you solve this particular \ph{unit}.
\par\medskip
Do not attempt the task now. Do not explain your choice.
\par\medskip
-{}-{}- BEGIN \ph{UNIT} -{}-{}-
\par
\ph{task description}
\par
-{}-{}- END \ph{UNIT} -{}-{}-
\par\medskip
Reply with exactly one line, nothing else:
\par
EFFORT: <one of: \ph{option 1}, \ldots, \ph{option $k$}>
\tabularnewline
\bottomrule
\end{tabular}

\vspace{0.75\baselineskip}

\begin{tabular}{@{}p{0.16\textwidth}p{0.78\textwidth}@{}}
\toprule
\multicolumn{2}{@{}l}{\textbf{(b) Effort options}} \\
\midrule
\textbf{Option} & \textbf{Description} \\
\midrule
\texttt{low} & the least thinking \textemdash{} fastest and cheapest \\
\texttt{medium} & moderate thinking \\
\texttt{high} & extended thinking \\
\texttt{xhigh}, \texttt{max} & the most thinking this model offers \textemdash{} slowest and dearest \\
\bottomrule
\end{tabular}

\vspace{0.75\baselineskip}

\begin{tabular*}{0.97\textwidth}{@{}l@{\extracolsep{\fill}}l p{0.62\textwidth}@{}}
\toprule
\multicolumn{3}{@{}l}{\textbf{(c) Benchmark-specific fields}} \\
\midrule
\textbf{Benchmark} & \textbf{Unit} & \textbf{Setting} \\
\midrule
Terminal-Bench 2.1 & task & You will solve a command-line task inside a Linux container, as an autonomous agent sending batches of shell keystrokes to a terminal and reading the screen back (the Terminus-2 scaffold). You get one attempt. You succeed only if the task's own verifier script passes afterwards. \\
\addlinespace
SWE-Bench Pro & instance & You will work on a real GitHub issue inside the repository's Docker container, as an autonomous agent driving a shell one command at a time (the mini-swe-agent scaffold). You get one attempt. You succeed only if the repository's own fail-to-pass and pass-to-pass tests both pass on the patch you leave behind. \\
\addlinespace
AIME 2026 & problem & You will answer this AIME competition problem in a single response, with no tools, no code execution and no second chance to revise. The answer is an integer between 0 and 999 and is marked right or wrong on that integer alone. \\
\addlinespace
GPQA Diamond & question & You will answer this graduate-level multiple-choice science question in a single response, with no tools, no search and no second chance to revise. Only the letter you pick is marked. \\
\bottomrule
\end{tabular*}
\end{table}

\section{Ablation: Repeated Agentic Runs}\label{sec:appendix-repeated-runs}

Our agentic benchmarks use one execution per task at each effort level, so individual task outcomes are subject to execution variability.
To assess whether our findings hold with repeated executions, we apply the same analyses to the official runs released by DeepSWE \citep{huang2026deepswe}.
DeepSWE v1.1 evaluates 113 long-horizon software-engineering tasks using \texttt{mini-swe-agent}, with four rollouts per task at each effort level.
The official release includes five models from our study: GPT-5.6-luna, GPT-5.6-terra, GPT-5.6-sol, and Claude Sonnet 5 at all settings from \texttt{low} to \texttt{max}, and GLM-5.2 at \texttt{high} and \texttt{max}.
We exclude DeepSeek V4 Flash because DeepSWE evaluates it only at \texttt{max}.
For this analysis, a task is considered solved at an effort level if at least one rollout passes; rollouts that terminate with an error are counted as failures.
We use the per-rollout results downloaded on September 23, 2026.
\subsection{Instances gained and lost}

Figure~\ref{fig:app_deepswe_instances_gained} extends the comparison in Section~\ref{sec:perf} to DeepSWE.
With four rollouts per task, a task solved at the lower effort level is considered lost at the higher level only if all four higher-effort rollouts fail.
Regressions persist under this stricter criterion: every pair of adjacent effort levels exhibits at least one lost task for every model.
For Claude Sonnet 5, increasing effort from \texttt{high} to \texttt{xhigh} gains 7 tasks but loses 12, reducing the number solved from 90 to 85 out of 113.
For GPT-5.6-sol, the same transition gains 2 tasks and loses 3.
GLM-5.2 gains 18 tasks and loses 8 when moving from \texttt{high} to \texttt{max}.
GPT-5.6-luna is the only model whose highest-effort solved set contains its lowest-effort solved set: \texttt{max} solves all five tasks solved at \texttt{low}.
The other four models lose between 2 and 11 tasks when moving from their lowest to their highest effort level.
Thus, increasing reasoning effort can still reduce task coverage when success is measured across four executions per task.

\begin{figure}[t]
    \centering
    \includegraphics[width=\linewidth]{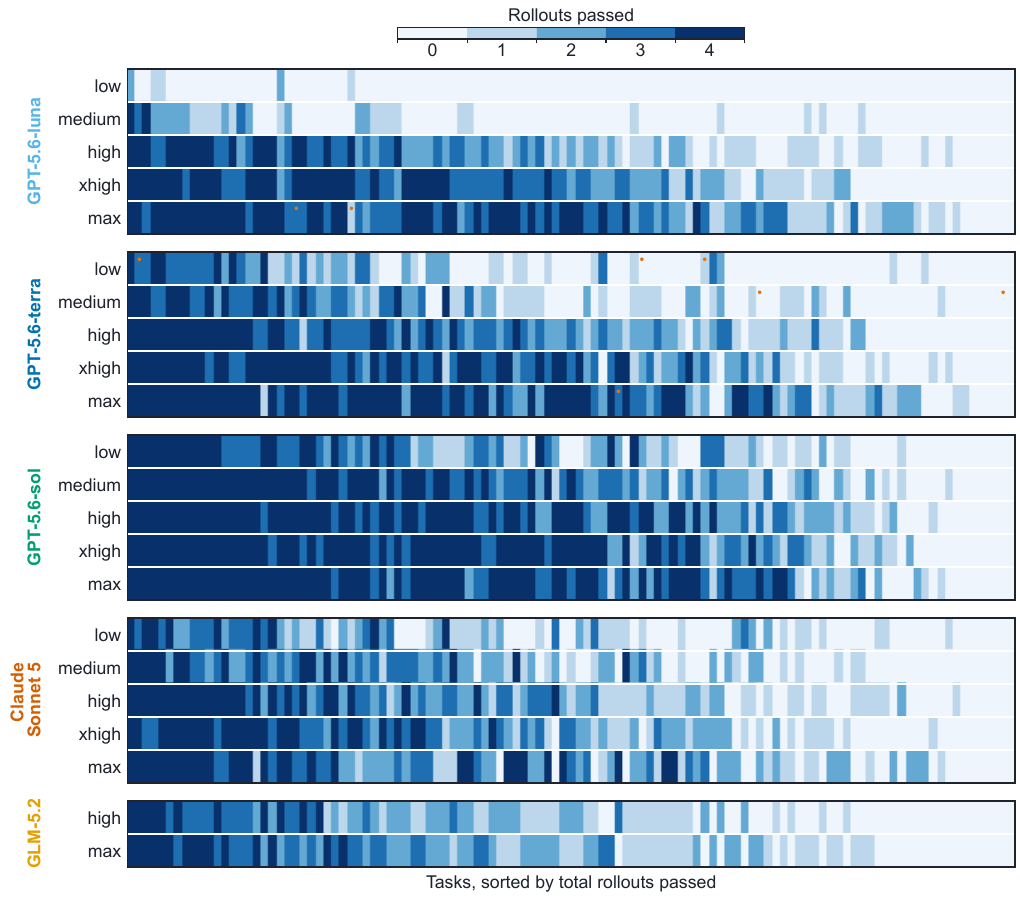}
    \caption{
    Rollouts passed on each DeepSWE task at every reasoning-effort setting.
    Each row is an effort setting and each column is a task; columns are sorted separately for each model by total rollouts passed, so a column position does not identify the same task across models.
    A task solved at a lower setting but not at a higher one appears as a shaded cell above an unshaded cell in the same column.
    Orange dots mark the eight cells with fewer than four successful rollouts in the official DeepSWE release.}
    \label{fig:app_deepswe_instances_gained}
\end{figure}

\subsection{Cost of always-solved tasks}

Figure~\ref{fig:app_deepswe_solvable_cost} repeats the always-solved cost comparison of Figure~\ref{fig:solvable_cost} on DeepSWE, and Figure~\ref{fig:app_deepswe_solvable_cost_full} adds the waiting cost of Appendix~\ref{sec:appendix-full-cost}.
DeepSWE does not run \texttt{none}, and we plot GPT-5.6 and Claude Sonnet 5 from \texttt{low} to \texttt{xhigh}, the highest level of our agentic runs, and GLM-5.2 at \texttt{high} and \texttt{max}.
We price each rollout from its token counts at the same rates as our own runs.
Claude Sonnet 5's trajectories do not record reasoning tokens, so its bars show input and output only.

The finding in Section~\ref{sec:cost} also holds on DeepSWE: among tasks solved at every plotted level, mean API cost increases with each successive effort level for all five models.
From \texttt{low} to \texttt{xhigh}, mean API cost rises by $4.16\times$ for GPT-5.6-sol, $4.45\times$ for GPT-5.6-terra, and $5.20\times$ for Claude Sonnet 5, compared with $2.51\times$ and $3.04\times$ for GPT-5.6-sol and Claude Sonnet 5 on Terminal-Bench.
GPT-5.6-luna's cost rises by $13.58\times$, although only 4 of the 113 tasks are solved at every level.
GLM-5.2's cost rises by $1.44\times$ from \texttt{high} to \texttt{max}.
As in our evaluations, input tokens account for most of the API cost, contributing 69--91\% of every bar, and reasoning accounts for at most 15\% of the runs.
Adding waiting cost preserves the ordering, and waiting accounts for 48--95\% of the full cost; from \texttt{low} to \texttt{xhigh}, full cost rises by $3.87\times$ for GPT-5.6-sol and $3.88\times$ for Claude Sonnet 5.

\begin{figure}[t]
    \centering
    \includegraphics[width=\linewidth]{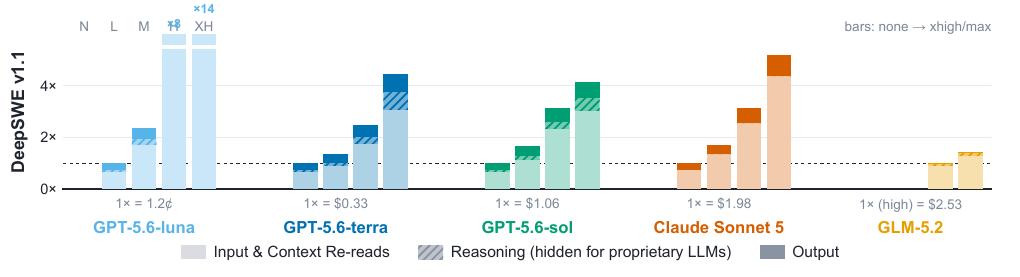}
    \caption{API cost of one DeepSWE task the model solves at every plotted effort level, relative to the cost at \texttt{low} (dashed line; GLM-5.2 is anchored on \texttt{high}).
    A task's cost at a level is the mean over all its passing rollouts.
    The footer gives the dollar value of $1\times$.
    Bars taller than the axis are clipped and labelled.}
    \label{fig:app_deepswe_solvable_cost}
\end{figure}

\begin{figure}[t]
    \centering
    \includegraphics[width=\linewidth]{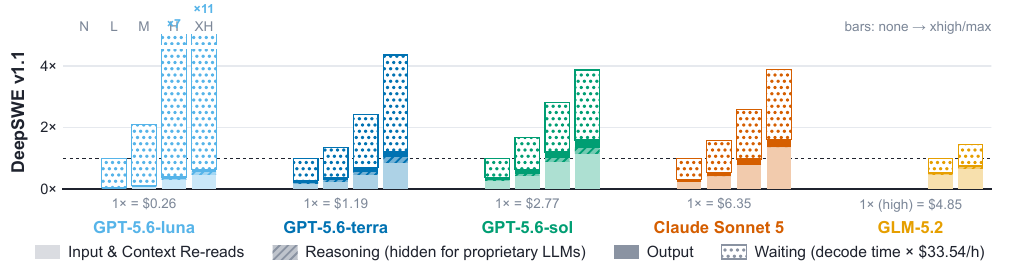}
    \caption{Full cost of one DeepSWE task the model solves at every plotted effort level, relative to the full cost at \texttt{low}.}
    \label{fig:app_deepswe_solvable_cost_full}
\end{figure}

\end{document}